\documentclass{article} % For LaTeX2e
\usepackage{iclr2027_conference,times}

\usepackage{amsmath,amsfonts,bm}

\def\eqref#1{equation~\ref{#1}}
\def\1{\bm{1}}

\DeclareMathAlphabet{\mathsfit}{\encodingdefault}{\sfdefault}{m}{sl}
\SetMathAlphabet{\mathsfit}{bold}{\encodingdefault}{\sfdefault}{bx}{n}

\usepackage[
    colorlinks=true,
    citecolor=blue,
    linkcolor=blue,
    urlcolor=blue
]{hyperref}

\usepackage{url}
\usepackage{verbatim}
\usepackage{booktabs}
\usepackage{algpseudocode}
\usepackage{algorithm}
\usepackage{multirow,multicol}
\usepackage{amsmath}
\usepackage{wrapfig}
\usepackage{bbm}
\usepackage{subcaption}
\usepackage{graphicx}
\usepackage[table]{xcolor}

\title{Autoregressive latent diffusion for 3D molecule generation}

\author{Federico Ottomano, Gaopeng Ren, Kim E. Jelfs, Alex M. Ganose \\%\thanks{Use footnote for providing further information
Department of Chemistry\\
Imperial College London \\
London, UK \\
\texttt{f.ottomano@imperial.ac.uk} \\
\And
Yingzhen Li \\
Department of Computing \\
Imperial College London \\
London, UK \\\\
}

\iclrfinalcopy % Uncomment for camera-ready version, but NOT for submission.
\begin{document}

\maketitle

\begin{abstract}
Three-dimensional (3D) molecule generation has been dominated by diffusion models, which achieve strong generation quality but typically require molecular size to be specified (or predicted) separately before generation. This can be limiting for fragment-based molecule generation, central to drug discovery, where the size of the generated structure is itself part of the design problem. Autoregressive models determine size during generation and naturally support partial-structure conditioning, but balancing unconditional and fragment-conditioned generation remains challenging.
We introduce KRONOS, a latent autoregressive diffusion framework that generates molecules in the latent space of a Unified AutoEncoder (UAE), jointly modeling molecular graph topology and geometry, while retaining the flexibility of autoregressive generation. We further introduce a mixed training strategy inspired by the Fill-in-the-Middle (FIM) paradigm, enabling a single left-to-right autoregressive model to support both unconditional and fragment-conditioned generation. Experiments on QM9 and GEOM-Drugs demonstrate strong unconditional generation performance and competitive fragment-conditioned generation.
\end{abstract}

\section{Introduction}
Generative modeling of three-dimensional (3D) molecules is a central problem in machine learning for drug discovery and molecular design. Recent progress has largely been driven by diffusion~\citep{hoogeboom2022edm, huang2024jodo, luo2025towards, joshi2025allatom} and flow matching~\citep{irwin2024semlaflow}, which achieve excellent generation quality but typically require molecular size to be specified or predicted separately before generation. This is particularly relevant for molecule generation from partial structures, where the size of the generated structure is itself part of the design problem.

Autoregressive models provide a complementary paradigm by generating molecules sequentially~\citep{gebauer2019symmetry, luo2022gsphere, daigavane2024symphony}, determining molecular size during generation while naturally supporting conditioning on partial molecular structures. This flexibility is particularly valuable in medicinal chemistry, where new molecules are often designed by elaborating known scaffolds or functional fragments~\citep{lim2020scaffold,erlanson2016twenty}. Recent autoregressive models~\citep{cheng2025quetzal, rose2025neat} have narrowed the performance gap to diffusion-based approaches, but balancing unconditional generation with partial-structure conditioning remains challenging.

Unified latent representations~\citep{joshi2025allatom, xu2023geometric, park2025guiding} provide a natural interface for generative modeling by jointly encoding discrete structural information and continuous geometry. Existing latent molecular generators, however, have primarily relied on diffusion-based generation.

In this work, we introduce \textbf{KRONOS}, a latent autoregressive diffusion framework that operates directly in the latent space of a Unified AutoEncoder (UAE)~\citep{luo2025towards}. KRONOS models latent molecular tokens autoregressively, while parameterizing each conditional distribution with a diffusion model~\citep{li2024mar}. An explicit stop-prediction mechanism enables variable-length generation. To support fragment-conditioned generation, we further introduce a mixed training strategy inspired by the Fill-in-the-Middle (FIM) paradigm~\citep{bavarian2022fim} enabling both unconditional and fragment-conditioned generation within a single left-to-right architecture.

Experiments on QM9 and GEOM-Drugs datasets demonstrate that KRONOS achieves leading unconditional performance among autoregressive methods, while remaining competitive with diffusion-based approaches. Furthermore, the results show that enabling fragment-conditioned generation has negligible impact on unconditional generation quality. A comprehensive discussion of related work is provided in Appendix~\ref{app:related_work}.
Our contributions are summarized below:
\begin{itemize}
    \item We introduce KRONOS, a latent autoregressive diffusion framework for 3D molecule generation.

    \item We introduce a fragment-conditioned training strategy that supports both unconditional and fragment-conditioned generation within a single autoregressive model.

    \item We demonstrate strong unconditional generation performance, together with competitive fragment-conditioned generation.
\end{itemize}

\section{Preliminaries}
\paragraph{3D molecule representation}
A 3D molecule consists of discrete chemical attributes (atom types and bond connectivity) together with continuous geometry (3D coordinates). This hybrid nature makes joint generative modeling challenging, as it requires simultaneously handling discrete and continuous variables while respecting permutation and SE(3) symmetries. Following previous work~\citep{luo2025towards, huang2024jodo}, we represent a 3D molecule as a tuple $\mathbf{G} = (\mathbf{H}, \mathbf{A}, \mathbf{X})$, where $\mathbf{H} \in \mathbb{R}^{|\mathcal{V}| \times d_1}$ encodes atomic features, $\mathbf{A} \in \mathbb{R}^{|\mathcal{V}| \times |\mathcal{V}| \times d_2}$ is the bond adjacency matrix, and $\mathbf{X} \in \mathbb{R}^{|\mathcal{V}| \times 3}$ specifies 3D atomic coordinates. $\mathcal{V}$ represents the set of atoms of $\mathbf{G}$, while $d_1$ and $d_2$ represent the dimension of atomic and bond features, respectively. 

\paragraph{Unified Autoencoder (UAE).}
To obtain a unified representation, we adopt the UAE framework~\citep{luo2025towards}, which encodes each atom into a continuous latent vector that jointly captures its chemical identity, bonding environment, and spatial position. The UAE encoder $\mathcal{E}_{\text{UAE}}$ employs a graph relational transformer~\citep{diao2023relational}, which maps a molecule $\mathbf{G}$ into a latent sequence $\mathcal{E}_{\text{UAE}}(\mathbf{G}) = \mathbf{Z} = (\mathbf{z}_1,\ldots,\mathbf{z}_n)$, where $\mathbf{z}_i \in \mathbb{R}^d$. A transformer decoder $\mathcal{D}_{\text{UAE}}$ decodes latents back into the reconstructed molecule $\hat{\mathbf{G}} = \mathcal{D}_{\text{UAE}}(\mathbf{Z})$.
Notably, both the encoder $\mathcal{E}_{\text{UAE}}$ and decoder $\mathcal{D}_{\text{UAE}}$ are permutation equivariant with respect to atom ordering. Furthermore, the UAE is trained with random SE(3) transformations of the input coordinates, encouraging the encoder–decoder pair to learn SE(3)-equivariant reconstruction.
%The encoder produces a sequence of latent tokens $\mathbf{Z} = (\mathbf{z}_1, \dots, \mathbf{z}_n)$, while a corresponding decoder reconstructs the full molecular structure from these latents. Crucially, both encoder and decoder are permutation equivariant, allowing generation to be defined over unordered sets of atoms.

\paragraph{Autoregressive Diffusion for Continuous Tokens.}
To model conditional distributions over continuous tokens, we adopt the
Diffusion Loss introduced by~\citet{li2024mar}, which provides a unified
framework for continuous-valued generation in both autoregressive
(e.g.,~\citet{vaswani2017attention}) and masked modeling settings
(e.g.,~\citet{devlin2019bert, chang2022maskgit}). This formulation replaces
discrete token prediction with a denoising objective, enabling flexible
modeling of continuous tokens without discretization.
In particular, each conditional distribution $p(\mathbf{z} \mid \mathbf{c})$ is parameterized
via a diffusion-based denoising process:
\begin{equation}
\mathcal{L}(\mathbf{c}, \mathbf{z})
=
\mathbb{E}_{t, \varepsilon}
\left[
\left\|
\varepsilon -
\varepsilon_\theta(\mathbf{z}_t \mid t, \mathbf{c})
\right\|^2
\right],
\end{equation}
where $\mathbf{z}_t$ is a noise-perturbed version of $\mathbf{z}$ and $\mathbf{c}$ represents the conditioning context produced by the autoregressive model. This formulation has been successfully applied to 3D molecule generation for modeling continuous 3D geometry~\citep{cheng2025quetzal}.

\section{Methods}
Prior autoregressive approaches to 3D molecule generation operate in data
space~\citep{cheng2025quetzal, rose2025neat}. %and predict discrete atom identities and continuous geometry in separate output spaces.
This requires the model to learn over heterogeneous discrete and continuous generation targets. %While effective, this requires the model to operate over heterogeneous discrete and continuous output spaces.
We instead perform autoregressive generation directly in the latent space of a UAE~\citep{luo2025towards}, where each latent token jointly encodes atom identity, bond topology, and 3D geometry. KRONOS combines three key components: (i) autoregressive modeling over latent 3D molecular tokens (Section \ref{sec:autoreg_latent}), (ii) a mixed autoregressive training strategy supporting both unconditional and fragment-conditioned generation (Section~\ref{sec:frag_cond_auto}) and (iii) a Diffusion Loss formulation~\citep{li2024mar} with a stop classifier for sequence termination (Section~\ref{sec:train_obj}). %Together, these components enable a unified framework for both unconditional and fragment-conditioned 3D molecule generation. 
\begin{figure}[t]
    \centering
    \includegraphics[width=1.00\columnwidth]{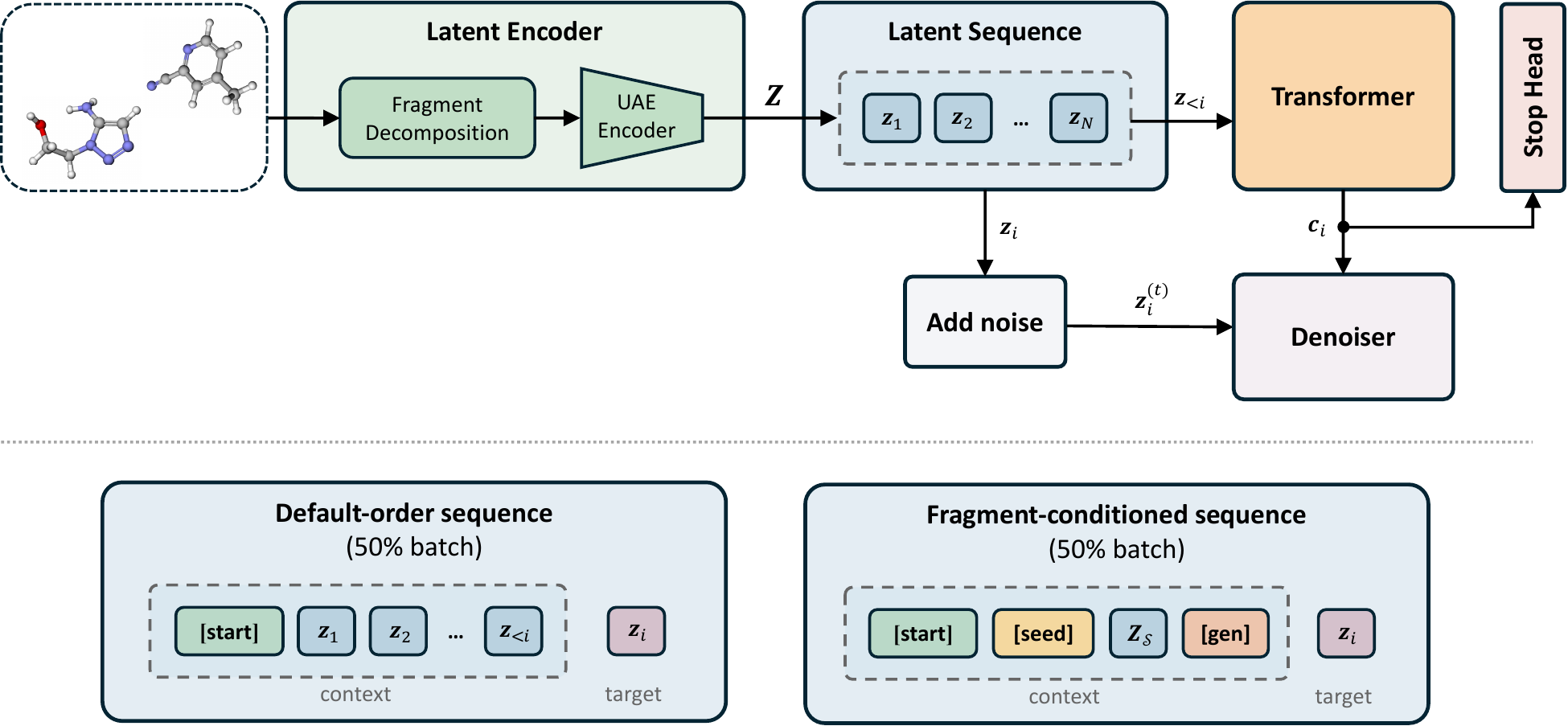}
    \captionsetup{skip=12pt}
    \caption{Visual overview of KRONOS. Molecules are decomposed into fragments and encoded into latent molecular sequences by the UAE encoder. During training, each example follows either the default-order or fragment-conditioned sequence construction. A causal transformer encodes the resulting autoregressive context, which conditions both a latent diffusion denoiser for next-token prediction and a stop head for sequence termination. This unified training strategy supports both unconditional and fragment-conditioned generation without architectural modifications.}
    \label{fig:framework}
    %\vspace{-10pt}
\end{figure}
We provide a visual overview of KRONOS in Figure~\ref{fig:framework}.

\subsection{Autoregressive Latent Generation}
\label{sec:autoreg_latent}
Our goal is to learn a generative model over latent molecular sequences
$p_\theta(\mathbf{z}_{1:n})$, where each latent token corresponds to an atom.
Samples from this distribution are decoded into molecular structures using
the UAE decoder $\mathcal{D}_{\text{UAE}}$.
We model the joint distribution over latent tokens
$p_\theta(\mathbf{z}_{1:n})$ using a left-to-right autoregressive factorisation:
\begin{equation}
    p_\theta(\mathbf{z}_{1:n})
    =
    \prod_{i=1}^{n}
    p_\theta(\mathbf{z}_i \mid \mathbf{z}_{<i}),
    \label{eq:lr_ar}
\end{equation}
where generation is initialized with a special $[\mathrm{START}]$ token.
At each step $i$, a causal transformer encodes the preceding tokens into a
context representation
\begin{equation}
    \mathbf{c}_i =
    \mathrm{Transformer}([\mathrm{START}], \mathbf{z}_1, \ldots, \mathbf{z}_{i-1}),
    \label{eq:transf_context}
\end{equation}
which conditions a diffusion model used to parameterize
$p_\theta(\mathbf{z}_i \mid \mathbf{z}_{<i})$ (described in Section~\ref{subsec:diffusion}). %described in Section~\ref{sec:diffusion}. 
As the latent space does not admit a natural end-of-sequence
token, we model sequence termination separately via a stop classifier
$p_\theta([\mathrm{STOP}] \mid \mathbf{c}_i)$ (described in Section~\ref{subsec:stop}).
During training, the autoregressive factorization in Eq.~\ref{eq:lr_ar} corresponds to the \emph{default-order} sequence, which follows the token ordering inherited from the input molecules, and is used for unconditional (\emph{de novo}) molecule generation. This ordering likely traces
back to how the 3D structures were initialized from SMILES~\citep{cheng2025quetzal}.

\subsection{Fragment-conditioned generation}
\label{sec:frag_cond_auto}
Previous autoregressive models for 3D molecule generation employ a fixed atom ordering~\citep{cheng2025quetzal}.
While effective for unconditional generation, training with a fixed atom ordering limits generalization to arbitrary subsets of molecular atoms.
To address this, we introduce a mixed autoregressive training strategy inspired by the Fill-in-the-Middle (FIM) paradigm~\citep{bavarian2022fim}, originally proposed to unlock document-infilling capabilities within left-to-right transformer models and later extended to protein language models~\citep{lee2023protfim}.
During training, each molecule is transformed into either a default-order sequence (Section~\ref{sec:autoreg_latent}) or a \textit{fragment-conditioned} sequence with probabilities $1-p_{f}$ and $p_{f}$, respectively, where $p_f$ denotes the fragment-conditioning probability.  We use $p_f = 0.5$ throughout the main experiments and study the effect of varying $p_f$ in Appendix~\ref{app:p_f}. The fragment-conditioned sequence construction can be viewed as defining autoregressive factorizations at the fragment level rather than at the atom level. We introduce this perspective below.

\paragraph{Permutation space}
By the chain rule, a joint distribution over $n$ latent variables can be factorized according to any permutation
$\pi\in S_n$:
\begin{equation}
    p_\theta(\mathbf{z}_{1:n})
    = \prod_{i=1}^{n} p_\theta\!\left(\mathbf{z}_{\pi(i)} \mid
          \mathbf{z}_{\pi(<i)} \right).
\end{equation}
The permutation equivariance of the UAE encoder–decoder pair ensures that different orderings of the same molecule correspond to consistently permuted latent sequences, which decode to the same molecular structure. This allows us to construct alternative autoregressive factorizations by choosing different molecular token orderings.
Order-agnostic autoregressive (OA-AR) models~\citep{uria2016neural,yang2019xlnet}
treat $\pi$ as a latent variable and maximise the expected log-likelihood over
$S_n$.  Because the full expectation is intractable, they optimise the
variational lower bound
\begin{equation}\label{eq:oaar_obj}
    \log p_\theta(\mathbf{z})
    \;\ge\;
    \mathbb{E}_{\pi\sim S_n}\!\left[\,
      \sum_{i=1}^{n}\log p_\theta\!\left(
        \mathbf{z}_{\pi(i)} \mid \mathbf{z}_{\pi(<i)}\right)
    \right],
\end{equation}
which can be estimated with Monte-Carlo sampling of permutations and positions, as originally proposed by~\citet{uria2016neural} and later reformulated to enable efficient, single-step training~\citep{hoogeboom2022ardm}. Rather than marginalizing over $S_n$ as in OA-AR, we instead define autoregressive factorisations based on \textit{molecular fragments}. Specifically, each molecule is decomposed into chemically meaningful fragments. A random subset of these fragments is then selected as the conditioning context, and a single deterministic ordering is assigned to the resulting conditioning and generation sequences. We examine the importance of preserving structured atom orderings in Appendix~\ref{app:perm-ablation}.

\paragraph{Fragment decomposition}
To construct fragment-conditioned training sequences, each
molecule is partitioned into $K$ fragments
$\{\mathcal F_0,\ldots,\mathcal F_{K-1}\}$ using a BRICS-based
decomposition~\citep{degen2008brics}, a common fragmentation strategy employed in molecular machine learning~\citep{zhang2021motifbased, wang2022improving, lee2026fragfm}. Hydrogens bonded to heteroatoms are retained within their parent fragment, as they can contribute to the fragment's chemical identity. Carbon-bound structural hydrogens are instead assigned to a dedicated pseudo-fragment $\mathcal{F}_H \notin \{\mathcal F_0,\ldots,\mathcal F_{K-1}\}$, as they follow from the heavy-atom skeleton. This yields a collection of contiguous fragment blocks used to construct fragment-based orderings. Full details are provided in Appendix~\ref{app:frag_decomp}.
\paragraph{Fragment-conditioned sequence}
Given the fragment decomposition above, for molecules with $K \ge 2$ we sample a seed size $s \sim \mathrm{Uniform}\{1,\ldots,K-1\}$ and sample a fragment subset
$\mathcal S \subset \{\mathcal F_0,\ldots,\mathcal F_{K-1}\}$ uniformly among all subsets of size $s$. For molecules with $K=1$, we instead use the default autoregressive sequence. The selected subset $\mathcal{S}$ serves as the \emph{seed}, while the
complement $\mathcal{S}^c$ defines the generation target. 
For a sampled subset $\mathcal S$, we construct
$\mathbf Z_{\mathcal S}$ and $\mathbf Z_{\mathcal S^c}$ as the latent token sequences associated with fragments in $\mathcal S$ and its complement $\mathcal S^c$, respectively. Fragments and their constituent atoms are ordered deterministically within each subset $\mathcal{S}$ and $\mathcal{S}^c$ (see Appendix~\ref{app:frag_decomp}). Importantly, the latent tokens associated with structural hydrogens $\mathbf{Z}_{\mathcal{F}_H}$ are excluded from the fragment sampling procedure. Instead, they are appended after $\mathbf{Z}_{\mathcal{S}^c}$ in every training sequence and generated autoregressively. The resulting training sequence is

\begin{equation}
[\mathrm{START}]\;
\circ
[\mathrm{SEED}]\;
\circ
\mathbf{Z}_{\mathcal{S}}\;
\circ
[\mathrm{GEN}]\;
\circ
\mathbf{Z}_{\mathcal{S}^c}\;
\circ
\mathbf{Z}_{\mathcal{F}_H}\;
\circ
[\mathrm{STOP}],
\end{equation}

where sentinel tokens mark the conditioning and generation regions, $\circ$ denotes sequence concatenation, and $[\mathrm{STOP}]$ delimits the end of the training sequence. Sentinel tokens are structural markers rather than generation targets: the diffusion objective is applied only at atom
positions. Conditioned on the seed sequence $\mathbf Z_{\mathcal S}$, the remaining tokens are generated autoregressively:
\begin{equation}
p_\theta(\tilde{\mathbf{Z}}_{\mathcal{S}^c} \mid \mathbf Z_{\mathcal S})
=
\prod_{j=1}^{m}
p_\theta(\mathbf z_{r_j}\mid \mathbf Z_{\mathcal S}, \mathbf z_{r_{<j}}),
\end{equation}
where $\tilde{\mathbf{Z}}_{\mathcal{S}^c} = \mathbf{Z}_{\mathcal{S}^c} \circ \mathbf{Z}_{\mathcal{F}_H} = (\mathbf{z}_{r_1}, \dots , \mathbf{z}_{r_m})$ denotes the concatenated target sequence.
Unlike conventional FIM, our formulation does not perform two-sided infilling between a prefix and suffix; instead, the FIM-style transformation constructs seed-conditioned prefix-completion tasks, where a sampled subset of fragments provides the context, and the remaining fragments are generated autoregressively. In contrast to order-agnostic autoregressive training~\citep{uria2016neural}, which marginalizes over token orderings, we use a single deterministic factorization for each sampled fragment partition, fixing the fragment and within-fragment atom orderings. Diversity across training factorizations therefore arises from sampling different seed subsets.

\subsection{Training objective}
\label{sec:train_obj}
\paragraph{Diffusion loss}
\label{subsec:diffusion}
To model autoregressive conditionals $p_\theta(\mathbf z_i \mid \mathbf z_{<i})$, 
we adopt a Diffusion Loss formulation~\citep{li2024mar} and model each using a diffusion process in the continuous latent space. Let $
\mathbf c_i$
denote the causal transformer context at step $i$. To model the distribution
of the next latent token $\mathbf z_i$, we adopt a Variance-Preserving (VP) forward process~\citep{song2021scorebased} that perturbs $\mathbf z_i$ with Gaussian noise:
\begin{equation}
\mathbf z_i^{(t)} =
\alpha_t \mathbf z_i +
\sigma_t \varepsilon,
\quad
t \sim \mathcal U[0,1],\;
\varepsilon \sim \mathcal N(0,I) \, ,
\end{equation}

where $\alpha_t$ and $\sigma_t$ denote the signal and noise coefficients of the VP noise schedule. The denoising network $\varepsilon_\theta(\cdot)$ is implemented as a residual Multi-Layer Perceptron (MLP) with Adaptive Layer Normalization (AdaLN) conditioning~\citep{li2024mar}. The network predicts the injected noise $\varepsilon$ conditioned on the noisy latent, the autoregressive context, and the diffusion timestep. Following standard diffusion training~\citep{ho2020ddpm}, we optimize
the noise prediction objective
\begin{equation}
\label{eq:diff_loss}
\mathcal L_{\mathrm{diff}}
=
\mathbb E_{(\mathbf z_i,\mathbf c_i),\,t,\,\varepsilon}
\left[
\left\|
\varepsilon_\theta(\mathbf z_i^{(t)}, \mathbf c_i, t)
-
\varepsilon
\right\|^2
\right] \ ,
\end{equation}
where $\mathbf c_i$ denotes the autoregressive transformer context (Eq.~\ref{eq:transf_context}), and $\mathbf z_i$ and $\mathbf z_i^{(t)}$ denote the clean and noisy latent targets, respectively.
\paragraph{Stop loss}
\label{subsec:stop}
Unlike autoregressive models operating directly in data space~\citep{cheng2025quetzal, rose2025neat}, our latent representations do not contain a dedicated end-of-sequence token. We therefore train a lightweight stop classifier $\mathrm{MLP}_{\mathrm{stop}}$ on the transformer hidden states $\mathbf c_i$. At each step, it predicts the probability that generation should terminate conditioned on the current context:
\begin{equation}
    p_\theta([\mathrm{STOP}] \mid \mathbf c_i)= \sigma\!\left( \mathrm{MLP}_{\mathrm{stop}}(\mathbf c_i)\right),
\end{equation}
where $\sigma(\cdot)$ denotes the sigmoid function. 
We train the classifier jointly with the diffusion model using binary cross-entropy
\begin{equation}
\label{eq:stop_loss}
\mathcal{L}_{\mathrm{stop}}
=
\mathbb{E}_{(\mathbf c_i, y_i^{\mathrm{stop}})}
\left[
\mathrm{BCE}\!\left(
p_\theta([\mathrm{STOP}] \mid \mathbf c_i),
y_i^{\mathrm{stop}}
\right)
\right] \, ,
\end{equation}
where $y_i^{\mathrm{stop}} \in \{0,1\}$ is one when $\mathbf c_i$ corresponds
to the context after the final latent token, and zero otherwise. At inference, generation stops when $\sigma(\text{MLP}_{\text{stop}}(\mathbf{c}_i)) > \tau$, with $\tau=0.5$. Additional details are provided in Appendix~\ref{app:sampling}.

\paragraph{Global objective} All components, the transformer, the MLP denoiser, 
and the stop classifier, are trained jointly, while the UAE 
encoder is kept frozen. The full objective is:
\begin{equation}
    \mathcal{L}
    = \mathcal{L}_{\mathrm{diff}} + \lambda_{\mathrm{stop}}\mathcal{L}_{\mathrm{stop}} \ ,
\end{equation}
where $\mathcal{L}_{\mathrm{diff}}$, $\mathcal{L}_{\text{stop}}$ are defined in Eq.~\ref{eq:diff_loss}, Eq.~\ref{eq:stop_loss}, respectively, and $\lambda_{\mathrm{stop}}$ 
is set to $1.0$.

\begin{table*}[t]
\centering
\small
\caption{
3D metrics on the QM9 dataset for the unconditional generation task.
\textbf{Best} and \underline{second-best} entries are highlighted in bold and underlined, respectively. Models marked with (*) were evaluated using released generated molecules.
Models marked with (**) were evaluated using official pre-trained weights and source code. Models marked with (\textdagger) were re-trained utilizing official source code. We report metrics evaluated on the test set of QM9 dataset in \textcolor{gray}{gray}.
For (**) and (\textdagger) we report $\text{mean} \pm \text{std}$ over three independent inference runs.}

\resizebox{\textwidth}{!}{
\begin{tabular}{lrrrrrr}
\toprule
\textbf{Model}

& \multicolumn{1}{c}{\textbf{3D Validity} ($\uparrow$)}

& \multicolumn{1}{c}{\textbf{3D AtomStable} ($\uparrow$)}

& \multicolumn{1}{c}{\textbf{3D MolStable} ($\uparrow$)}

& \multicolumn{1}{c}{\textbf{Bond length} ($\downarrow$)}

& \multicolumn{1}{c}{\textbf{Bond angle} ($\downarrow$)}

& \multicolumn{1}{c}{\textbf{Dihedral angle} ($\downarrow$)} \\

& \multicolumn{1}{c}{(\%)}
& \multicolumn{1}{c}{(\%)}
& \multicolumn{1}{c}{(\%)}
& \multicolumn{1}{c}{$(\times 10^{-2})$}
& \multicolumn{1}{c}{$(\times 10^{-2})$}
& \multicolumn{1}{c}{$(\times 10^{-3})$} \\

\midrule

\textcolor{gray}{QM9}

& \textcolor{gray}{$98.4\phantom{_{\pm 0.0}}$}
& \textcolor{gray}{$99.53\phantom{_{\pm 0.00}}$}
& \textcolor{gray}{$96.3\phantom{_{\pm 0.0}}$}
& \textcolor{gray}{$0.003\phantom{_{\pm}}$}
& \textcolor{gray}{$1.06\phantom{_{\pm 0.00}}$}
& \textcolor{gray}{$0.05\phantom{_{\pm 0.00}}$} \\

\midrule
\multicolumn{7}{l}{\textit{Diffusion / Flow}}\\
\midrule
ADiT*
& $79.3\phantom{_{\pm 0.0}}$
& $76.43\phantom{_{\pm 0.00}}$
& $7.6\phantom{_{\pm 0.0}}$
& $111.5\phantom{_{\pm 0.0}}$
& $14.90\phantom{_{\pm 0.00}}$
& $3.60\phantom{_{\pm 0.00}}$ \\

JODO*
& $\underline{96.3}\phantom{_{\pm 0.0}}$
& $\mathbf{99.25}\phantom{_{\pm 0.00}}$
& $\mathbf{93.6}\phantom{_{\pm 0.0}}$
& $38.1\phantom{_{\pm 0.0}}$
& $2.23\phantom{_{\pm 0.00}}$
& $1.37\phantom{_{\pm 0.00}}$ \\

EDM\textsuperscript{\textdagger}
& $95.3_{\pm0.1}$
& $97.56_{\pm0.19}$
& $87.6_{\pm0.1}$
& $23.7_{\pm2.3}$
& $2.21_{\pm0.06}$
& $\underline{0.50}_{\pm0.10}$ \\

GeoLDM**
& $93.3_{\pm0.2}$
& $98.88_{\pm0.03}$
& $89.5_{\pm0.1}$
& $42.8_{\pm0.6}$
& $2.24_{\pm0.06}$
& $0.56_{\pm0.04}$ \\

SemlaFlow**
& $85.8_{\pm0.1}$
& $97.07_{\pm0.03}$
& $78.6_{\pm0.2}$
& $75.2_{\pm0.8}$
& $1.94_{\pm0.02}$
& $1.55_{\pm0.21}$ \\

MiDi**
& $95.3_{\pm 0.3}$
& $98.21_{\pm 0.05}$
& $83.5_{\pm 0.2}$
& $110.5_{\pm 1.7}$
& $3.00_{\pm 0.10}$
& $0.80_{\pm 0.10}$ \\

\midrule
\multicolumn{7}{l}{\textit{Autoregressive}}\\
\midrule

Symphony*
& $68.1\phantom{_{\pm 0.0}}$
& $90.75\phantom{_{\pm 0.00}}$
& $43.9\phantom{_{\pm 0.0}}$
& $\underline{19.4}\phantom{_{\pm 0.0}}$
& $11.00\phantom{_{\pm 0.00}}$
& $2.06\phantom{_{\pm 0.00}}$ \\

G-SchNet*
& $80.1\phantom{_{\pm 0.0}}$
& $93.71\phantom{_{\pm 0.00}}$
& $63.7\phantom{_{\pm 0.0}}$
& $41.5\phantom{_{\pm 0.0}}$
& $9.25\phantom{_{\pm 0.00}}$
& $6.49\phantom{_{\pm 0.00}}$ \\

G-SphereNet*
& $16.4\phantom{_{\pm 0.0}}$
& $67.82\phantom{_{\pm 0.00}}$
& $14.0\phantom{_{\pm 0.0}}$
& $27.8\phantom{_{\pm 0.0}}$
& $37.60\phantom{_{\pm 0.00}}$
& $14.30\phantom{_{\pm 0.00}}$ \\

%UDM-3D
%& $92.5 \pm 0.2$
%& $97.58 \pm 0.04$
%& $76.3 \pm 0.4$
%& $\underline{6.1 \pm 0.4}$
%& $\mathbf{1.30 \pm 0.10}$
%& $1.10 \pm 0.20$ \\

QUETZAL**
& $94.8_{\pm 0.1}$
& $98.28_{\pm 0.04}$
& $87.4_{\pm 0.2}$
& $40.1_{\pm 1.1}$
& $2.20_{\pm 0.10}$
& $\mathbf{0.41}_{\pm 0.03}$ \\

NEAT**
& $93.4_{\pm 0.1}$
& $98.09_{\pm 0.02}$
& $84.0_{\pm 0.3}$
& $33.1_{\pm 0.1}$
& $\underline{1.90}_{\pm 0.10}$
& $4.30_{\pm 0.20}$ \\

\midrule 

\rowcolor{blue!15}
\textbf{KRONOS (ours)}
& $\mathbf{97.3}_{\pm 0.1}$
& $\underline{99.07}_{\pm 0.02}$
& $\underline{92.6}_{\pm 0.1}$
& $\mathbf{4.8}_{\pm 0.6}$
& $\mathbf{1.33}_{\pm 0.03}$
& $1.00_{\pm 0.10}$ \\
\bottomrule
\end{tabular}
}
\label{table:metrics_qm9_3d}
\end{table*}

\begin{table*}[t]
\centering
\small
\caption{
3D metrics on the GEOM-Drugs dataset for the unconditional generation task.
\textbf{Best} and \underline{second-best} entries are highlighted in bold and underlined, respectively. We report metrics evaluated on the test set of GEOM-Drugs dataset in \textcolor{gray}{gray}. Models marked with (*) were evaluated using released generated molecules.
Models marked with (**) were evaluated using official pre-trained weights and source code. Models marked with (\textdagger) were re-trained utilizing the official source code. For (**) and (\textdagger) we report $\text{mean} \pm \text{std}$ over three independent inference runs.
}

\resizebox{\textwidth}{!}{
\begin{tabular}{lccccccc}
\toprule
\textbf{Model}
& \textbf{3D Validity} ($\uparrow$)
& \textbf{3D AtomStable} ($\uparrow$)
& \textbf{3D MolStable} ($\uparrow$)
& \textbf{Bond length} ($\downarrow$)
& \textbf{Bond angle} ($\downarrow$)
& \textbf{Dihedral angle} ($\downarrow$) \\

& \multicolumn{1}{c}{(\%)}
& \multicolumn{1}{c}{(\%)}
& \multicolumn{1}{c}{(\%)}
& 
&
&
\\
\midrule
\textcolor{gray}{GEOM}
& \textcolor{gray}{$96.6\phantom{_{\pm 0.0}}$} & \textcolor{gray}{$86.17\phantom{_{\pm 0.00}}$} & \textcolor{gray}{$2.80\phantom{_{\pm 0.00}}$} & \textcolor{gray}{$0.253\phantom{_{\pm 0.000}}$} & \textcolor{gray}{$0.2695\phantom{_{\pm 0.0000}}$} & \textcolor{gray}{$0.0153\phantom{_{\pm 0.0000}}$} \\
%UDM-3D
%& $/$ & $/$ & $/$ & $/$ & $/$ & $/$ \\
\midrule
\multicolumn{7}{l}{\textit{Diffusion / Flow}}\\
\midrule
ADiT*
& $97.3\phantom{_{\pm 0.0}}$
& $75.46\phantom{_{\pm 0.00}}$
& $0.12\phantom{_{\pm 0.00}}$
& $1.323\phantom{_{\pm 0.000}}$
& $1.0474\phantom{_{\pm 0.0000}}$
& $0.0548\phantom{_{\pm 0.0000}}$ \\

JODO*
& $95.6\phantom{_{\pm 0.0}}$
& $84.49\phantom{_{\pm 0.00}}$
& $1.11\phantom{_{\pm 0.00}}$
& $\underline{0.089}\phantom{_{\pm 0.000}}$
& ${\underline{0.0099}\phantom{_{\pm 0.0000}}}$
& $\mathbf{0.0005\phantom{_{\pm 0.0000}}}$
\\

EDM\textsuperscript{\textdagger}
& $90.5_{\pm 0.6}$
& $83.35_{\pm 0.05}$
& $1.01_{\pm0.04}$
& $0.865_{\pm0.005}$
& $0.8368_{\pm0.0060}$
& $0.0567_{\pm0.0004}$
\\

GeoLDM**
& $\mathbf{99.4}_{\pm 0.1}$
& $84.30_{\pm 0.04}$
& $0.59_{\pm0.02}$
& $0.916_{\pm0.006}$
& $0.7885_{\pm0.0024}$
& $0.0474_{\pm0.0003}$
\\

SemlaFlow**
& $92.4_{\pm 0.2}$
& $84.10_{\pm 0.09}$
& $\underline{2.17}_{\pm0.24}$
& $0.569_{\pm0.001}$
& $0.1241_{\pm0.0015}$
& $\underline{0.0007}_{\pm0.0000}$
\\

MiDi**
& $92.0_{\pm0.4}$
& $75.48_{\pm0.16}$
& $0.30_{\pm0.04}$
& $0.113_{\pm0.001}$
& $0.0915_{\pm0.0029}$
& $0.0046_{\pm0.0003}$\\

\midrule
\multicolumn{7}{l}{\textit{Autoregressive}}\\
\midrule

QUETZAL\textsuperscript{\textdagger}
& $93.8_{\pm 0.2}$
& $\underline{86.27}_{\pm 0.03}$
& $2.04_{\pm 0.06}$
& $0.882_{\pm 0.008}$
& $0.9370_{\pm 0.0068}$
& $0.0614_{\pm 0.0015}$ \\

NEAT**
& $\underline{98.1}_{\pm 0.1}$
& $81.16_{\pm 0.04}$
& $0.50_{\pm 0.01}$
& $1.194_{\pm 0.002}$
& $1.3939_{\pm 0.0342}$
& $0.1543_{\pm 0.0012}$ \\

\midrule
\rowcolor{blue!15}
\textbf{KRONOS (ours)}
& $94.7_{\pm0.2}$ & $\mathbf{86.45}_{\pm0.03}$ & $\mathbf{2.70}_{\pm0.10}$ & $\mathbf{0.048}_{\pm0.003}$ & $\mathbf{0.0094}_{\pm0.0004}$ & $\mathbf{0.0005}_{\pm0.0002}$ \\

\bottomrule
\end{tabular}
}
\label{table:metrics_geom_3d}
\end{table*}

\section{Experiments}
We evaluate KRONOS on three complementary tasks: unconditional 3D molecule generation on the QM9 and GEOM-Drugs datasets (Tables~\ref{table:metrics_qm9_3d} and~\ref{table:metrics_geom_3d}), fragment-conditioned generation from partial molecular context (Table~\ref{tab:pref_cond}), and conformational stability assessed through semi-empirical GFN2-xTB optimization (Figure~\ref{fig:xtb_eval}). 
The UAE is trained from scratch with inter-fragment message passing disabled in the encoder to ensure fragment-local latent representations (see Appendix~\ref{app:frag_local}).
Additional experiments on 2D molecule generation are provided in Appendix~\ref{app:2d_gen}.

\subsection{Experimental setup}
\paragraph{Datasets} We evaluate our framework on QM9~\citep{ramakrishnan2014qm9} and GEOM-Drugs~\citep{axelrod2022geom} datasets. For QM9, we use the same train/validation/test split established in existing work~\citep{luo2025towards,huang2024jodo}, resulting in $97,609$/$14,405$/$12,405$ molecules. Similarly, for GEOM-Drugs we follow the pre-processing and train/validation/test splits utilized in existing literature~\citep{vignac2023midi}, retaining the five lowest-energy conformers per molecule. This results in $1,172,532$/$146,405$/$146,793$ conformers for training, validation, and testing, respectively.
\paragraph{Baselines}
We compare the proposed method with several existing autoregressive and diffusion baselines for 3D molecule generation. Autoregressive baselines include G-SchNet~\citep{gebauer2019symmetry}, G-SphereNet~\citep{luo2022gsphere}, Symphony~\citep{daigavane2024symphony}, QUETZAL~\citep{cheng2025quetzal}, NEAT~\citep{rose2025neat}. Diffusion models include EDM~\citep{hoogeboom2022edm}, MiDi~\citep{vignac2023midi}, ADiT~\citep{joshi2025allatom}, GeoLDM~\citep{xu2023geometric} and JODO~\citep{huang2024jodo}. SemlaFlow~\citep{irwin2024semlaflow} is included as a flow matching baseline. We do not report performance of UDM-3D~\citep{luo2025towards}, because we are unable to reproduce reasonable results utilizing the source code, and pre-trained checkpoints or pre-generated molecules are unavailable at the time of writing.
\paragraph{Metrics} We evaluate generated molecules using standard metrics for 3D molecule generation. Specifically, we report validity and stability of reconstructed molecular graphs together with \textit{Maximum Mean Discrepancy} (MMD)~\citep{gretton2012mmd} between generated and reference bond length, bond angle, and dihedral angle distributions. Complete metric definitions are provided in Appendix~\ref{app:eval_protocol}.

\subsection{Results}
\paragraph{Unconditional molecule generation} Results on QM9 and GEOM-Drugs are reported in Tables~\ref{table:metrics_qm9_3d} and~\ref{table:metrics_geom_3d}, respectively.
On QM9, KRONOS achieves the highest validity and the strongest overall performance among autoregressive methods, while remaining competitive with diffusion-based models. As shown in Appendix~\ref{app:p_f}, reducing the fragment-conditioning probability $p_f$ further improves unconditional generation performance, indicating that the small performance trade-off primarily arises from the mixed training strategy. On GEOM-Drugs, KRONOS demonstrates strong unconditional generation performance across the evaluated metrics. Compared with recent autoregressive baselines such as QUETZAL and NEAT, these results demonstrate that latent autoregressive generation can improve both molecular stability and geometric fidelity. 

\paragraph{Assessment of conformational stability}
While the molecular stability criterion (\textbf{3D MolStable}) introduced by~\citet{hoogeboom2022edm} is a standard evaluation metric for molecule generation, it relies on heuristic bond-assignment procedures that can be sensitive to aromatic systems and distance thresholds~\citep{rose2025neat, daigavane2024symphony, irwin2024semlaflow}. %Consequently, it primarily evaluates whether a generated atomic point cloud admits a chemically plausible molecular graph, rather than directly assessing the physical plausibility of the generated conformation. 
\begin{figure}[t]
\centering
\includegraphics[width=0.88\textwidth]{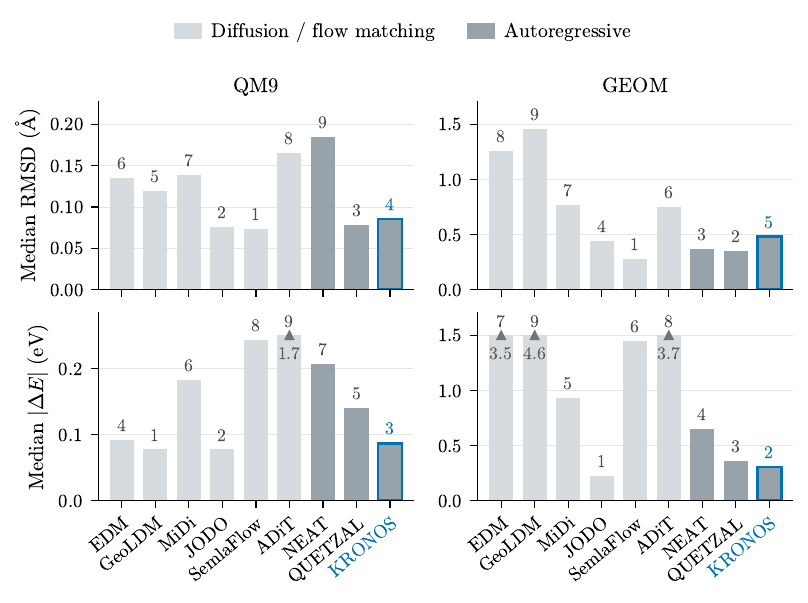}
\caption{
Physical plausibility of generated conformers. We report the median RMSD between generated and xTB-relaxed geometries (top) and the median absolute energy difference between them (bottom). Numbers above the bars indicate rank within each panel ($1$ = best); bars marked with a caret exceed the axis limit and show their true value.
}
\label{fig:xtb_eval}
\end{figure}
To complement this graph-based evaluation, we perform semi-empirical quantum-chemical refinement using GFN2-xTB~\citep{bannwarth2019xtb}. For each generated molecule, we compute its single-point energy before optimization and report (i) the Root Mean Square Deviation (RMSD) between the generated and optimized conformations after optimal rigid-body alignment using the Kabsch algorithm~\citep{kabsch1976solution}, and (ii) the absolute energy difference $|\Delta E|$. Lower values indicate closer agreement with the optimized structure. We evaluate $10{,}000$ generated molecules per model. Figure~\ref{fig:xtb_eval} summarizes the results.
On QM9, KRONOS attains the third-lowest median energy difference (within $0.01$~eV of the best model), together with a competitive median RMSD. On GEOM-Drugs, it attains the second-lowest median energy difference, while geometric agreement with the optimized structures is more variable. Overall, the results indicate consistent energetic agreement with the xTB-optimized structures, whereas geometric agreement is more variable across datasets.

\begin{figure}[t]
\centering
\includegraphics[width=0.90\textwidth]{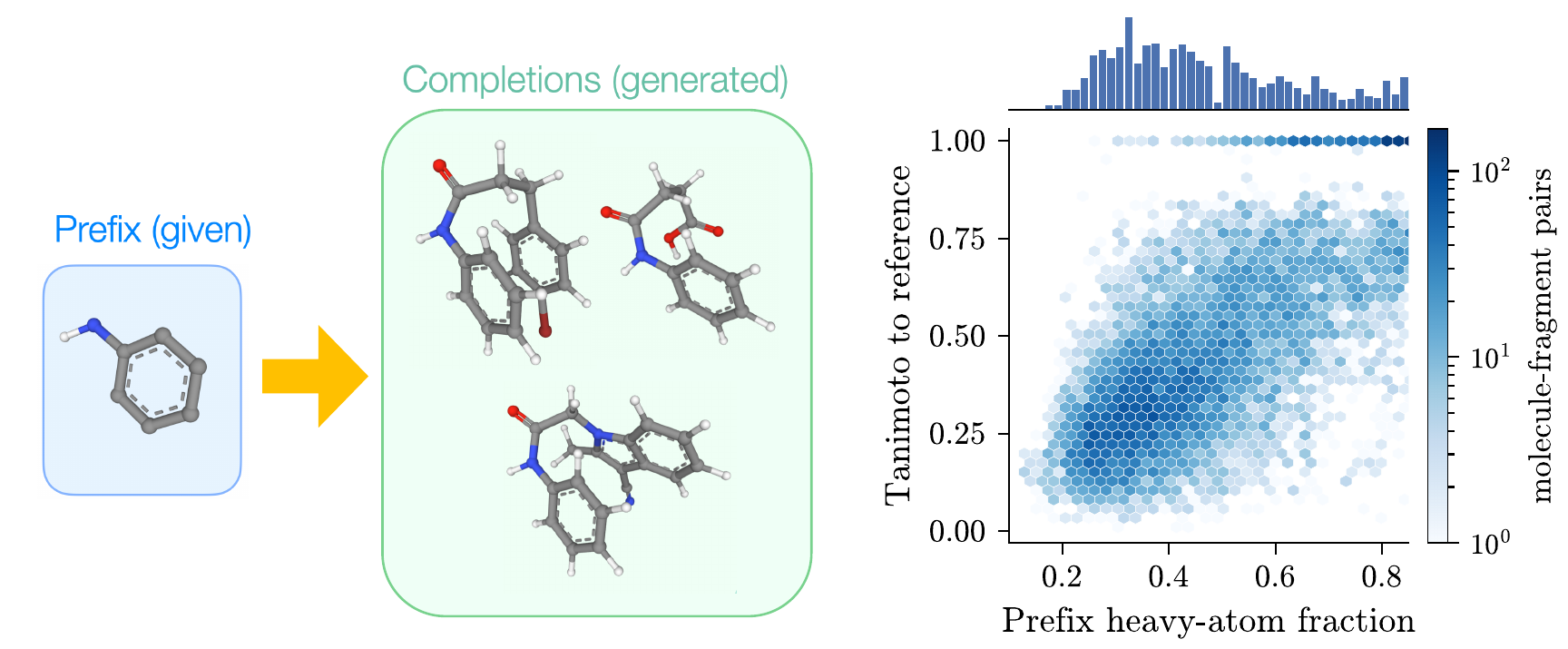}
\caption{Fragment-conditioned molecule generation. \textit{Left}) a conditioning fragment (``prefix'') extracted from a molecule in the GEOM-Drugs test set, and three independent completions generated by KRONOS, that extend the fragment into diverse, chemically plausible 3D molecules. \textit{Right}) prefix heavy-atom fraction versus Tanimoto similarity to the \emph{reference} (the
parent molecule the prefix was taken from).}
\label{fig:mol_completions}
\end{figure}

\paragraph{Fragment-conditioned molecule generation}
We evaluate whether the proposed framework can generate chemically coherent 3D molecules from partial molecular context. This task follows the spirit of \emph{prefix completion} introduced by~\citet{rose2025neat}. 
However, rather than conditioning on a fixed set of manually designed scaffolds, we construct conditioning prefixes by our fragment decomposition scheme (Appendix~\ref{app:frag_decomp}) applied to molecules from the GEOM-Drugs test set, and treat each resulting fragment as conditioning context. 
\begin{table*}[!h]
\centering
\small
\caption{
Fragment-conditioned molecule generation performance over $\sim15$k molecule-fragment pairs. Models marked with (**) were evaluated using official pre-trained weights and source code. Models marked with (\textdagger) were re-trained utilizing the official source code. \textbf{Best} and \underline{second-best} entries are highlighted in bold and underlined, respectively.
Results are aggregated across molecule--fragment pairs and reported as mean $\pm$ 95\% confidence interval.
}
\label{tab:pref_cond}
\setlength{\tabcolsep}{10pt}

\resizebox{\textwidth}{!}{
\begin{tabular}{lccc|ccc|c}
\toprule

&
\multicolumn{3}{c|}{\textbf{Mean@3}} &
\multicolumn{3}{c|}{\textbf{Best@3}} &
\textbf{Unique@3}
\\

\cmidrule(lr){2-4}
\cmidrule(lr){5-7}

\textbf{Model}
& 
\textbf{3D Validity}
&
\textbf{3D AtomStable}
&
\textbf{3D MolStable}
&
\textbf{3D Validity}
&
\textbf{3D AtomStable}
&
\textbf{3D MolStable}
&

\\

\midrule

QUETZAL\textsuperscript{\textdagger}
&
$48.7_{\pm0.6}$
&
$67.6_{\pm0.2}$
&
$0.2_{\pm0.1}$
&
$70.6_{\pm0.7}$
&
$74.2_{\pm0.2}$
&
$0.5_{\pm0.1}$
&
$41.0_{\pm 0.5}$
\\

NEAT**
&
$\mathbf{93.7}_{\pm0.3}$
&
$\mathbf{83.4}_{\pm0.1}$
&
$\underline{0.8}_{\pm0.1}$
&
$\mathbf{97.4}_{\pm0.2}$
&
$\mathbf{87.9}_{\pm0.1}$
&
$\mathbf{2.1}_{\pm0.2}$
&
$\mathbf{93.7}_{\pm 0.3}$
\\

\midrule

\rowcolor{blue!15}
\textbf{KRONOS} (\textbf{ours})
&
$\underline{79.7}_{\pm0.4}$
&
$\underline{82.1}_{\pm0.1}$
&
$\mathbf{0.9}_{\pm0.1}$
&
$\underline{96.2}_{\pm0.3}$
&
$\underline{86.5}_{\pm0.1}$
&
$\underline{1.8}_{\pm0.2}$
&
$\underline{74.2}_{\pm 0.4}$
\\

\bottomrule
\end{tabular}
}
\end{table*}
The same fragment-derived prefixes are used to evaluate KRONOS, NEAT, and QUETZAL. 
This setting represents a practical, conditional-design scenario in which a fragment from a known three-dimensional structure is retained while alternative molecular structures are generated around it, as arises in scaffold-based design~\citep{lim2020scaffold}; conditioning on a fragment resolved by X-ray crystallography and elaborated from its bound pose is likewise standard in fragment-based drug discovery~\citep{erlanson2016twenty}. %Accordingly, each prefix retains the atomic coordinates of the fragment in its parent conformer.
We collect $\sim15$k molecule-fragment pairs and sample $3$ completions per pair. We report Best@3, Mean@3 and Unique@3 in Table~\ref{tab:pref_cond}; Mean@3 averages each metric across the three completions, Best@3 selects the best value, and Unique@3 measures the fraction of valid completions that are distinct based on their inferred 2D molecular graphs. Figure~\ref{fig:mol_completions} (left) illustrates the task.

The results show that fragment-conditioned generation can be learned without a fully permutation-invariant model. NEAT achieves higher overall metrics on this task, consistent with its set-based, permutation-invariant formulation and training objective, in which every example is a partial-structure completion task. KRONOS nevertheless achieves competitive fragment-conditioned generation while retaining strong unconditional generation performance (Tables~\ref{table:metrics_qm9_3d}, \ref{table:metrics_geom_3d}) and efficient sampling~(Appendix~\ref{app:runtime}). It substantially improves over QUETZAL, which shares its ordered autoregressive formulation but is trained only on fixed left-to-right sequences.

To assess how the amount of conditioning structure affects molecular exploration, we measure the similarity of generated completions to the parent molecule from which each conditioning fragment was extracted. Similarity is computed on 2D graphs obtained by perceiving bond orders from the generated coordinates using \texttt{rdDetermineBonds} from \texttt{RDKit}~\citep{rdkit}; the distance-based inference used for 3D metrics assign bond orders, but for GEOM-Drugs these are reduced to single bonds when constructing the molecular graph, following established conventions~\cite{hoogeboom2022edm, rose2025neat}; the resulting SMILES therefore cannot distinguish, e.g., aromatic from saturated rings. For each molecule-fragment pair, we retain the completion with highest Tanimoto similarity to the reference. When at most half of the reference's heavy atoms are supplied ($63\%$ of pairs), the median similarity is $0.32$ and the retained completion is identical to the reference molecule in only $0.4\%$ of pairs. Similarity increases with the size of the conditioning fragment, with reconstruction becoming common only once more than $60\%$ of the reference heavy atoms are supplied (Figure~\ref{fig:mol_completions} (right)). The size of the conditioning fragment therefore acts as a control on exploration: small fragments yield diverse elaborations around a retained core, while larger fragments increasingly determine the resulting molecule. Appendix~\ref{app:gen_mols} (Figure~\ref{fig:prefix_completion_examples}) shows representative completions. %in which an initial fragment is extended into distinct ring systems and substitution patterns.
%Completions spread around this value while their Tanimoto similarity to the reference never exceeds $0.5$, showing that KRONOS produces molecules of reasonable size without reconstructing the reference .
\section{Conclusion}
We introduced KRONOS, a latent autoregressive diffusion framework for 3D molecule generation. By operating in the unified latent space of a UAE, KRONOS generates molecular topology and geometry through a single, continuous token representation while its autoregressive formulation naturally supports variable-length generation. A FIM-inspired, fragment-conditioned training strategy further enables unconditional and fragment-conditioned generation within a single architecture. Experiments on QM9 and GEOM-Drugs demonstrate strong unconditional generation performance and competitive fragment-conditioned generation, while semi-empirical geometry optimization supports the physical plausibility of generated conformations. Overall, our results establish latent autoregressive generation as a flexible pathway for 3D molecule design. We discuss limitations and directions for future work in Appendix~\ref{app:lim_and_future}.

\begin{comment}
\section*{AI Disclosure}
In this work, we used generative AI tools to assist with research execution,
including adapting existing baseline implementations and accelerating
experimental workflows. AI tools were also used for literature search and to polish and improve the readability of the manuscript. We did not use generative AI tools to design the core methodology or experiments. All AI-assisted work was reviewed and verified by the authors. In particular, AI-assisted code and analyses were checked for correctness, while AI-assisted text was reviewed and edited for accuracy, consistency, and readability. The authors take full responsibility for the final content of this work, including all claims, analyses, code, and other artifacts.
\end{comment}
\section{Acknowledgments}
F.O., K.E.J., A.M.G., Y.Li acknowledge the AI for Chemistry: AIchemy hub for funding (EPSRC grant EP/Y028775/1 and EP/Y028759/1). G.R. acknowledges Imperial College London for funding through President's PhD Scholarships. The authors acknowledge the use of resources provided by the Isambard-AI National AI Research Resource (AIRR). Isambard-AI is operated by the University of Bristol and is funded by the UK Government’s Department for Science, Innovation and Technology (DSIT) via UK Research and Innovation; and the Science and Technology Facilities Council [ST/AIRR/I-A-I/1023].

\section*{Code Availability Statement}
Code, including instructions for reproducing all experiments, is available at \url{https://github.com/fedeotto/kronos}.

\bibliography{iclr2027_conference}

@InProceedings{hoogeboom2022edm,
  title = 	 {Equivariant Diffusion for Molecule Generation in 3{D}},
  author =       {Hoogeboom, Emiel and Satorras, Victor Garcia and Vignac, Cl{\'e}ment and Welling, Max},
  booktitle = 	 {Proceedings of the 39th International Conference on Machine Learning},
  pages = 	 {8867--8887},
  year = 	 {2022},
  editor = 	 {Chaudhuri, Kamalika and Jegelka, Stefanie and Song, Le and Szepesvari, Csaba and Niu, Gang and Sabato, Sivan},
  volume = 	 {162},
  series = 	 {Proceedings of Machine Learning Research},
  month = 	 {17--23 Jul},
  publisher =    {PMLR},
  url = 	 {https://proceedings.mlr.press/v162/hoogeboom22a.html}
}

@inproceedings{
irwin2024semlaflow,
title={Efficient \text{3D} Molecular Generation with Flow Matching and Scale Optimal Transport},
author={Ross Irwin and Alessandro Tibo and Jon Paul Janet and Simon Olsson},
booktitle={ICML 2024 AI for Science Workshop},
year={2024},
url={https://openreview.net/forum?id=CxAjGjdkqu}
}

@article{gebauer2019symmetry,
  title={Symmetry-adapted generation of 3d point sets for the targeted discovery of molecules},
  author={Gebauer, Niklas and Gastegger, Michael and Sch{\"u}tt, Kristof},
  journal={Advances in neural information processing systems},
  volume={32},
  year={2019}
}

@inproceedings{daigavane2024symphony,
title={Symphony: Symmetry-Equivariant Point-Centered Spherical Harmonics for 3\text{D} Molecule Generation},
author={Ameya Daigavane and Song Eun Kim and Mario Geiger and Tess Smidt},
booktitle={The Twelfth International Conference on Learning Representations},
year={2024},
url={https://openreview.net/forum?id=MIEnYtlGyv}
}

@inproceedings{
song2021scorebased,
title={Score-Based Generative Modeling through Stochastic Differential Equations},
author={Yang Song and Jascha Sohl-Dickstein and Diederik P Kingma and Abhishek Kumar and Stefano Ermon and Ben Poole},
booktitle={International Conference on Learning Representations},
year={2021},
url={https://openreview.net/forum?id=PxTIG12RRHS}
}

@article{bannwarth2019xtb,
author = {Bannwarth, Christoph and Ehlert, Sebastian and Grimme, Stefan},
title = {\text{GFN2-xTB}—An Accurate and Broadly Parametrized Self-Consistent Tight-Binding Quantum Chemical Method with Multipole Electrostatics and Density-Dependent Dispersion Contributions},
journal = {Journal of Chemical Theory and Computation},
volume = {15},
number = {3},
pages = {1652-1671},
year = {2019},
doi = {10.1021/acs.jctc.8b01176},
note ={PMID: 30741547},
URL = { 
        https://doi.org/10.1021/acs.jctc.8b01176
},
eprint = { 
        https://doi.org/10.1021/acs.jctc.8b01176
}
}

@article{park2025guiding,
  title={Guiding Generative Models to Uncover Diverse and Novel Crystals via Reinforcement Learning},
  author={Park, Hyunsoo and Walsh, Aron},
  journal={arXiv preprint arXiv:2511.07158},
  year={2025}
}

@article{gretton2012mmd,
  author  = {Arthur Gretton and Karsten M. Borgwardt and Malte J. Rasch and Bernhard Sch{{\"o}}lkopf and Alexander Smola},
  title   = {A Kernel Two-Sample Test},
  journal = {Journal of Machine Learning Research},
  year    = {2012},
  volume  = {13},
  number  = {25},
  pages   = {723-773},
  url     = {http://jmlr.org/papers/v13/gretton12a.html}
}

@inproceedings{joshi2025allatom,
title={All-atom Diffusion Transformers: Unified generative modelling of molecules and materials},
author={Chaitanya K. Joshi and Xiang Fu and Yi-Lun Liao and Vahe Gharakhanyan and Benjamin Kurt Miller and Anuroop Sriram and Zachary Ward Ulissi},
booktitle={Forty-second International Conference on Machine Learning},
year={2025},
url={https://openreview.net/forum?id=89QPmZjIhv}
}

@article{kabsch1976solution,
  title={A solution for the best rotation to relate two sets of vectors},
  author={Kabsch, Wolfgang},
  journal={Acta Crystallographica Section A},
  volume={32},
  number={5},
  pages={922--923},
  year={1976}
}

@inproceedings{ho2020ddpm,
 author = {Ho, Jonathan and Jain, Ajay and Abbeel, Pieter},
 booktitle = {Advances in Neural Information Processing Systems},
 editor = {H. Larochelle and M. Ranzato and R. Hadsell and M.F. Balcan and H. Lin},
 pages = {6840--6851},
 publisher = {Curran Associates, Inc.},
 title = {Denoising Diffusion Probabilistic Models},
 url = {https://proceedings.neurips.cc/paper_files/paper/2020/file/4c5bcfec8584af0d967f1ab10179ca4b-Paper.pdf},
 volume = {33},
 year = {2020}
}

@inproceedings{
karras2022elucidating,
title={Elucidating the Design Space of Diffusion-Based Generative Models},
author={Tero Karras and Miika Aittala and Timo Aila and Samuli Laine},
booktitle={Advances in Neural Information Processing Systems},
editor={Alice H. Oh and Alekh Agarwal and Danielle Belgrave and Kyunghyun Cho},
year={2022},
url={https://openreview.net/forum?id=k7FuTOWMOc7}
}

@inproceedings{
song2021denoising,
title={Denoising Diffusion Implicit Models},
author={Jiaming Song and Chenlin Meng and Stefano Ermon},
booktitle={International Conference on Learning Representations},
year={2021},
url={https://openreview.net/forum?id=St1giarCHLP}
}

@article{erlanson2016twenty,
  title={Twenty years on: the impact of fragments on drug discovery},
  author={Erlanson, Daniel A and Fesik, Stephen W and Hubbard, Roderick E and Jahnke, Wolfgang and Jhoti, Harren},
  journal={Nature reviews Drug discovery},
  volume={15},
  number={9},
  pages={605--619},
  year={2016},
  publisher={Nature Publishing Group UK London}
}

@inproceedings{luo2022gsphere,
title={An \text{A}utoregressive \text{F}low \text{M}odel for \text{3D} \text{M}olecular \text{G}eometry \text{G}eneration from \text{S}cratch},
author={Youzhi Luo and Shuiwang Ji},
booktitle={International Conference on Learning Representations},
year={2022},
url={https://openreview.net/forum?id=C03Ajc-NS5W}
}

@misc{rdkit,
  title  = {{RDKit}: Open-source cheminformatics},
  author = {Landrum, Gregory},
  url    = {https://www.rdkit.org},
}

@inproceedings{vignac2023midi,
title={\text{MiDi}: \text{M}ixed \text{G}raph and \text{3D} \text{D}enoising \text{D}iffusion for \text{M}olecule \text{G}eneration},
author={Clement Vignac and Nagham Osman and Laura Toni and Pascal Frossard},
booktitle={ICLR 2023 - Machine Learning for Drug Discovery workshop},
year={2023},
url={https://openreview.net/forum?id=M6Ifac3G4HK}
}

@inproceedings{
liu2026graph,
title={Graph Diffusion Transformers are In-Context Molecular Designers},
author={Gang Liu and Jie Chen and Yihan Zhu and Michael Sun and Tengfei Luo and Nitesh V Chawla and Meng Jiang},
booktitle={The Fourteenth International Conference on Learning Representations},
year={2026},
url={https://openreview.net/forum?id=lJ87GN5zJc}
}

@misc{lee2026fragfm,
      title={Frag\text{FM}: Hierarchical Framework for Efficient Molecule Generation via Fragment-Level Discrete Flow Matching}, 
      author={Joongwon Lee and Seonghwan Kim and Seokhyun Moon and Hyunwoo Kim and Woo Youn Kim},
      year={2026},
      eprint={2502.15805},
      archivePrefix={arXiv},
      primaryClass={cs.LG},
      url={https://arxiv.org/abs/2502.15805}, 
}

@article{degen2008brics,
author = {Degen, Jörg and Wegscheid-Gerlach, Christof and Zaliani, Andrea and Rarey, Matthias},
title = {On the Art of Compiling and Using 'Drug-Like' Chemical Fragment Spaces},
journal = {ChemMedChem},
volume = {3},
number = {10},
pages = {1503-1507},
doi = {https://doi.org/10.1002/cmdc.200800178},
url = {https://chemistry-europe.onlinelibrary.wiley.com/doi/abs/10.1002/cmdc.200800178},
eprint = {https://chemistry-europe.onlinelibrary.wiley.com/doi/pdf/10.1002/cmdc.200800178},
year = {2008}
}

@misc{bavarian2022fim,
      title={Efficient Training of Language Models to \text{F}ill in the \text{M}iddle}, 
      author={Mohammad Bavarian and Heewoo Jun and Nikolas Tezak and John Schulman and Christine McLeavey and Jerry Tworek and Mark Chen},
      year={2022},
      eprint={2207.14255},
      archivePrefix={arXiv},
      primaryClass={cs.CL},
      url={https://arxiv.org/abs/2207.14255}, 
}

@article{huang2024jodo,
  author={Huang, Han and Sun, Leilei and Du, Bowen and Lv, Weifeng},
  journal={IEEE Transactions on Neural Networks and Learning Systems}, 
  title={Learning Joint \text{2-D} and \text{3-D} Graph Diffusion Models for Complete Molecule Generation}, 
  year={2024},
  volume={35},
  number={9},
  pages={11857-11871},
  doi={10.1109/TNNLS.2024.3416328}}

@inproceedings{xu2023geometric,
  title={Geometric latent diffusion models for 3d molecule generation},
  author={Xu, Minkai and Powers, Alexander S and Dror, Ron O and Ermon, Stefano and Leskovec, Jure},
  booktitle={International Conference on Machine Learning},
  pages={38592--38610},
  year={2023},
  organization={PMLR}
}

@article{lim2020scaffold,
    author = {Lim, Jaechang and Hwang, Sang-Yeon and Moon, Seokhyun and Kim, Seungsu and Kim, Woo Youn},
    title = {Scaffold-based molecular design with a graph generative model},
    journal = {Chemical Science},
    volume = {11},
    number = {4},
    pages = {1153-1164},
    year = {2020},
    month = {01},
    issn = {2041-6520},
    doi = {10.1039/c9sc04503a},
    url = {https://doi.org/10.1039/c9sc04503a},
    eprint = {https://pubs.rsc.org/sc/article-pdf/11/4/1153/6908695/c9sc04503a.pdf},
}

@inproceedings{hoogeboom2022ardm,
  title = {Autoregressive Diffusion Models},
  author = {Hoogeboom, Emiel and Gritsenko, Alexey A. and Bastings, Jasmijn and Poole, Ben and van den Berg, Rianne and Salimans, Tim},
  booktitle = {International Conference on Learning Representations (ICLR)},
  year = {2022}
}

@inproceedings{yang2019xlnet,
 author = {Yang, Zhilin and Dai, Zihang and Yang, Yiming and Carbonell, Jaime and Salakhutdinov, Russ R and Le, Quoc V},
 booktitle = {Advances in Neural Information Processing Systems},
 editor = {H. Wallach and H. Larochelle and A. Beygelzimer and F. d\textquotesingle Alch\'{e}-Buc and E. Fox and R. Garnett},
 pages = {},
 publisher = {Curran Associates, Inc.},
 title = {XLNet: Generalized Autoregressive Pretraining for Language Understanding},
 url = {https://proceedings.neurips.cc/paper_files/paper/2019/file/dc6a7e655d7e5840e66733e9ee67cc69-Paper.pdf},
 volume = {32},
 year = {2019}
}

@article{uria2016neural,
author = {Uria, Benigno and C\^{o}t\'{e}, Marc-Alexandre and Gregor, Karol and Murray, Iain and Larochelle, Hugo},
title = {Neural autoregressive distribution estimation},
year = {2016},
issue_date = {January 2016},
publisher = {JMLR.org},
volume = {17},
number = {1},
issn = {1532-4435},
journal = {J. Mach. Learn. Res.},
month = jan,
pages = {7184–7220},
numpages = {37}
}

@article{rose2025neat,
  title={\text{NEAT}: \text{N}eighborhood-\text{G}uided, \text{E}fficient, \text{A}utoregressive \text{S}et \text{T}ransformer for \text{3D} \text{M}olecular \text{G}eneration},
  author={Rose, Daniel and Jacob, Roxane Axel and Kirchmair, Johannes and Langer, Thierry},
  journal={arXiv preprint arXiv:2512.05844},
  year={2025}
}

@inproceedings{chang2022maskgit,
  title={Mask\text{GIT}: Masked \text{g}enerative \text{I}mage \text{T}ransformer},
  author={Chang, Huiwen and Zhang, Han and Jiang, Lu and Liu, Ce and Freeman, William T},
  booktitle={Proceedings of the IEEE/CVF conference on computer vision and pattern recognition},
  pages={11315--11325},
  year={2022}
}

@inproceedings{kazemnejad2023length,
author = {Kazemnejad, Amirhossein and Padhi, Inkit and Ramamurthy, Karthikeyan Natesan and Das, Payel and Reddy, Siva},
title = {The impact of positional encoding on length generalization in transformers},
year = {2023},
publisher = {Curran Associates Inc.},
address = {Red Hook, NY, USA},
booktitle = {Proceedings of the 37th International Conference on Neural Information Processing Systems},
articleno = {1082},
numpages = {37},
location = {New Orleans, LA, USA},
series = {NIPS '23}
}

@inproceedings{devlin2019bert,
  title={\text{BERT}: Pre-training of deep bidirectional transformers for language understanding},
  author={Devlin, Jacob and Chang, Ming-Wei and Lee, Kenton and Toutanova, Kristina},
  booktitle={Proceedings of the 2019 conference of the North American chapter of the association for computational linguistics: human language technologies, volume 1 (long and short papers)},
  pages={4171--4186},
  year={2019}
}

@article {axelrod2022geom,
	Title = {\text{GEOM}, energy-annotated molecular conformations for property prediction and molecular generation},
	Author = {Axelrod, Simon and Gómez-Bombarelli, Rafael},
	DOI = {10.1038/s41597-022-01288-4},
	Number = {1},
	Volume = {9},
	Month = {April},
	Year = {2022},
	Journal = {Scientific data},
	ISSN = {2052-4463},
	Pages = {185},
	URL = {https://europepmc.org/articles/PMC9023519},
}

@article{ramakrishnan2014qm9,
	author = {Ramakrishnan, Raghunathan and Dral, Pavlo O. and Rupp, Matthias and von Lilienfeld, O. Anatole},
	date = {2014/08/05},
	doi = {10.1038/sdata.2014.22},
	id = {Ramakrishnan2014},
	isbn = {2052-4463},
	journal = {Scientific Data},
	number = {1},
	pages = {140022},
	title = {Quantum chemistry structures and properties of 134 kilo molecules},
	url = {https://doi.org/10.1038/sdata.2014.22},
	volume = {1},
	year = {2014}}

@inproceedings{li2024mar,
  title     = {Autoregressive Image Generation without Vector Quantization},
  author    = {Li, Tianhong and Tian, Yonglong and Li, He and Deng, Mingyang and He, Kaiming},
  booktitle = {Advances in Neural Information Processing Systems},
  year      = {2024}
}

@article{igashov2024equivariant,
  title={Equivariant \text{3D}-conditional diffusion model for molecular linker design},
  author={Igashov, Ilia and St{\"a}rk, Hannes and Vignac, Cl{\'e}ment and Schneuing, Arne and Satorras, Victor Garcia and Frossard, Pascal and Welling, Max and Bronstein, Michael and Correia, Bruno},
  journal={Nature Machine Intelligence},
  volume={6},
  number={4},
  pages={417--427},
  year={2024},
  publisher={Nature Publishing Group UK London}
}

@inproceedings{
luo2025towards,
title={Towards Unified and Lossless Latent Space for 3D Molecular Latent Diffusion Modeling},
author={Yanchen Luo and Zhiyuan Liu and Yi Zhao and Sihang Li and Hengxing Cai and Kenji Kawaguchi and Tat-Seng Chua and Yang Zhang and Xiang Wang},
booktitle={The Thirty-ninth Annual Conference on Neural Information Processing Systems},
year={2025},
url={https://openreview.net/forum?id=g2XE40zTrj}
}

@inproceedings{
diao2023relational,
title={Relational Attention: Generalizing Transformers for Graph-Structured Tasks},
author={Cameron Diao and Ricky Loynd},
booktitle={The Eleventh International Conference on Learning Representations },
year={2023},
url={https://openreview.net/forum?id=cFuMmbWiN6}
}

@inproceedings{vaswani2017attention,
 author = {Vaswani, Ashish and Shazeer, Noam and Parmar, Niki and Uszkoreit, Jakob and Jones, Llion and Gomez, Aidan N and Kaiser, \L ukasz and Polosukhin, Illia},
 booktitle = {Advances in Neural Information Processing Systems},
 editor = {I. Guyon and U. Von Luxburg and S. Bengio and H. Wallach and R. Fergus and S. Vishwanathan and R. Garnett},
 pages = {},
 title = {Attention is All you Need},
 year = {2017}
}

@article{wang2022improving,
  title={Improving molecular contrastive learning via faulty negative mitigation and decomposed fragment contrast},
  author={Wang, Yuyang and Magar, Rishikesh and Liang, Chen and Barati Farimani, Amir},
  journal={Journal of Chemical Information and Modeling},
  volume={62},
  number={11},
  pages={2713--2725},
  year={2022},
  publisher={ACS Publications}
}

@article{lee2023protfim,
  title={Prot\text{FIM}: Fill-in-\text{M}iddle protein sequence design via protein language models},
  author={Lee, Youhan and Yu, Hasun},
  journal={arXiv preprint arXiv:2303.16452},
  year={2023}
}

@inproceedings{haviv2022transformer,
    title = "Transformer Language Models without Positional Encodings Still Learn Positional Information",
    author = "Haviv, Adi  and
      Ram, Ori  and
      Press, Ofir  and
      Izsak, Peter  and
      Levy, Omer",
    editor = "Goldberg, Yoav  and
      Kozareva, Zornitsa  and
      Zhang, Yue",
    booktitle = "Findings of the Association for Computational Linguistics: EMNLP 2022",
    month = dec,
    year = "2022",
    address = "Abu Dhabi, United Arab Emirates",
    publisher = "Association for Computational Linguistics",
    url = "https://aclanthology.org/2022.findings-emnlp.99/",
    doi = "10.18653/v1/2022.findings-emnlp.99",
    pages = "1382--1390"
}

@inproceedings{
zhang2021motifbased,
title={Motif-based Graph Self-Supervised Learning for Molecular Property Prediction},
author={Zaixi Zhang and Qi Liu and Hao Wang and Chengqiang Lu and Chee-Kong Lee},
booktitle={Advances in Neural Information Processing Systems},
editor={A. Beygelzimer and Y. Dauphin and P. Liang and J. Wortman Vaughan},
year={2021},
url={https://openreview.net/forum?id=gwGYN1fQY8H}
}

@misc{cheng2025quetzal,
      title={Scalable \text{A}utoregressive \text{3D} \text{M}olecule \text{G}eneration}, 
      author={Austin H. Cheng and Chong Sun and Alán Aspuru-Guzik},
      year={2025},
      eprint={2505.13791},
      archivePrefix={arXiv},
      primaryClass={cs.LG},
      url={https://arxiv.org/abs/2505.13791}, 
}
\bibliographystyle{iclr2027_conference}

\appendix
\section{Appendix}
\subsection{Related work}\label{app:related_work}
\paragraph{Diffusion-based models for 3D molecule generation} Diffusion models have become the dominant paradigm for 3D molecule generation, achieving state-of-the-art performance across standard benchmarks. Early approaches such as EDM~\citep{hoogeboom2022edm} introduced E(3)-equivariant diffusion models that jointly denoise atom coordinates and atom types to generate 3D molecular structures. MiDi~\citep{vignac2023midi} extends this framework by jointly generating molecular graph topology and 3D geometry. JODO~\citep{huang2024jodo} further develops unified joint 2D/3D diffusion through a dedicated graph transformer that explicitly models interactions between molecular topology and geometry. 
GeoLDM~\citep{xu2023geometric} introduced latent diffusion for 3D molecule generation, encoding molecular structures into continuous latent representations with invariant and equivariant features before applying diffusion in the latent space. 
More recently, ADiT~\citep{joshi2025allatom} explored latent diffusion over unified all-atom representations, enabling a single transformer-based model to generate both molecules and materials within a shared latent space. Beyond unconditional generation, diffusion has also been adapted to conditional design tasks such as linker generation~\citep{igashov2024equivariant}, where a connector is
generated between two given fragments, and its size must be supplied or produced by a separately trained predictor. Parallel to diffusion, flow-matching approaches such as SemlaFlow~\citep{irwin2024semlaflow} learned continuous transport maps from noise to molecular distributions, achieving competitive generation quality while requiring significantly fewer sampling steps. However, in both unconditional and conditional settings the molecular size is fixed or predicted
before generation; in the conditional case this limits the ability to adapt the generated structure's size to the given context.

\paragraph{Autoregressive models for 3D molecule generation}
Early autoregressive approaches generated molecules sequentially by predicting one atom at a time together with its 3D placement~\citep{gebauer2019symmetry, luo2022gsphere}. Symphony~\citep{daigavane2024symphony} subsequently demonstrated that incorporating higher-order E(3)-equivariant representations substantially improves autoregressive molecule generation, narrowing the gap to diffusion-based methods.
More recent autoregressive approaches, including QUETZAL~\citep{cheng2025quetzal} and NEAT~\citep{rose2025neat}, combine autoregressive generation with continuous generative models for molecular geometry, using diffusion and flow-matching objectives, respectively. These methods substantially narrow the performance gap to diffusion models while adaptively determining molecular size during generation. However, existing autoregressive models still face a performance trade-off between unconditional generation and flexible conditioning on partial molecular structures. Our work addresses this trade-off through a unified autoregressive training framework that jointly supports both objectives.
\newpage
\subsection{Fragment-based molecular decomposition}\label{app:frag_decomp}
\begin{figure}[!h]
    \centering
\includegraphics[width=0.98\textwidth]{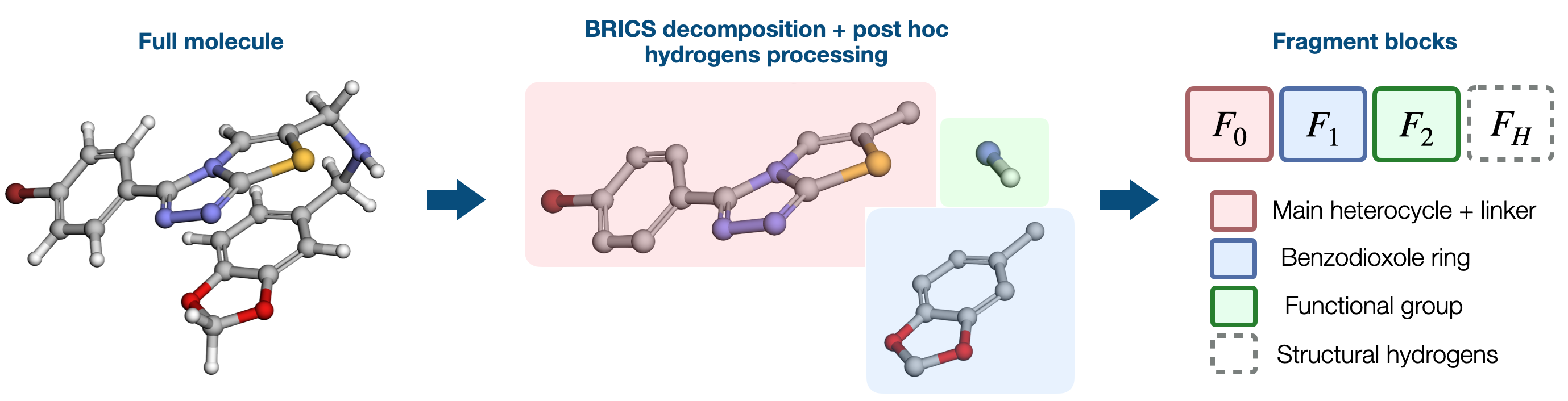}
    \caption{Visual example of the adopted molecular decomposition strategy. The initial molecule is first decomposed utilizing the BRICS algorithm. Then, hydrogens bonded to heteroatoms are retained by their parent fragment (preserving common functional groups as coherent units); structural hydrogens (carbon-bonded) are collected into a separate, pseudo-fragment that is always appended at the end of the obtained sequence.}
    \label{fig:frag_decomp}
    %\vspace{-10pt}
\end{figure}
Rather than considering arbitrary atom permutations, we decompose each molecule into chemically meaningful fragments and assign a deterministic ordering within each fragment. 
During training, random fragment partitions define the conditioning and generation regions, yielding diverse fragment-conditioned autoregressive generation tasks while avoiding arbitrary atom-level permutations. Fragment boundaries are obtained from a BRICS-based
decomposition~\citep{degen2008brics}, which cleaves bonds at retrosynthetically meaningful sites to produce chemically coherent building blocks, and has been explored in molecular generative modeling~\citep{lee2026fragfm, liu2026graph}. 
Removing these bonds partitions the molecular graph into connected components. Hydrogens bonded to heteroatoms are assigned to the same fragment as their parent atom, preserving functional groups such as $\mathrm{-OH}$, $\mathrm{-NH_2}$, and $\mathrm{-SH}$ as coherent units. In contrast, carbon-bound structural hydrogens are collected into a separate pseudo-fragment $\mathcal F_H$, which is always appended after all fragment blocks.
\begin{comment}
We further refine this decomposition by retaining heteroatom-bound hydrogens, such as hydroxyl or amine hydrogens,
within their parent fragment, while carbon-bound structural hydrogens are assigned to a separate pseudo-fragment. This preserves pharmacophoric functional
groups as coherent units.
Let $\mathcal V^{\text{heavy}}$ denote the heavy-atom set of a molecule. The BRICS-based
decomposition induces a partition
$\mathcal V^H=\mathcal F_0 \sqcup \cdots \sqcup \mathcal F_{K-1}$. Hydrogens are then incorporated according to chemical role. Hydrogens bonded to
heteroatoms are assigned to the same fragment as their parent atom, preserving functional groups such as $\mathrm{-OH}$, $\mathrm{-NH_2}$, and $\mathrm{-SH}$ as single units. 
In contrast, hydrogens bonded to carbon are collected into a separate pseudo-fragment $\mathcal F_H$, which is always appended after all
fragment blocks. Carbon-bound hydrogens are largely determined by the heavy-atom scaffold and contribute little independent structural information.
\end{comment}
Within each fragment $\mathcal F_k$, atoms follow a fixed ordering obtained through a deterministic depth-first search (DFS) traversal of the fragment subgraph.
%Within each fragment $\mathcal F_k$, atoms are ordered using a deterministic DFS traversal of the fragment subgraph, following prior work on DFS-based linearizations for autoregressive graph generation~\citep{cohen-karlik2023order}. 
%This fixes the internal order of each fragment and ensures that, within a fragment block, atoms are generated along a connected traversal of the
%fragment subgraph. Thus, each fragment contributes a contiguous block of atoms to the autoregressive sequence; permutations are applied only to the order of these
%blocks, not to the atoms within them.
%This defines the fragment-ordering set $\Pi_{\mathrm{frag}}$:
%each element of $\Pi_{\mathrm{frag}}$ is obtained by choosing a permutation of the $K$ (non-hydrogen) fragment blocks, keeping the DFS order fixed within each
%block, and appending $\mathcal F_H$ at the end. Hence
%$|\Pi_{\mathrm{frag}}| = K!$, which is dramatically smaller than $n!$. %while
%remaining chemically structured. 
%During training, the fragment set is randomly partitioned into conditioning and generation fragments while preserving the fixed DFS ordering within each fragment and the fragment ordering induced by the BRICS decomposition. Structural hydrogens are always appended last. Consequently, the model is exposed to many fragment-conditioned generation tasks while avoiding arbitrary atom-level permutations.
During training, the fragment set is randomly partitioned into conditioning and generation fragments. Within each set, fragments follow a fixed ordering, while atoms within each fragment follow the deterministic DFS ordering defined above. Structural hydrogens are always appended last. Consequently, the model is exposed to many fragment-conditioned generation tasks while avoiding arbitrary atom-level permutations.
We provide an illustrative example of the adopted decomposition in Figure~\ref{fig:frag_decomp}.

\subsection{Fragment-local latent encoding}\label{app:frag_local}
\begin{figure}[!h]
    \centering
\includegraphics[width=0.98\textwidth]{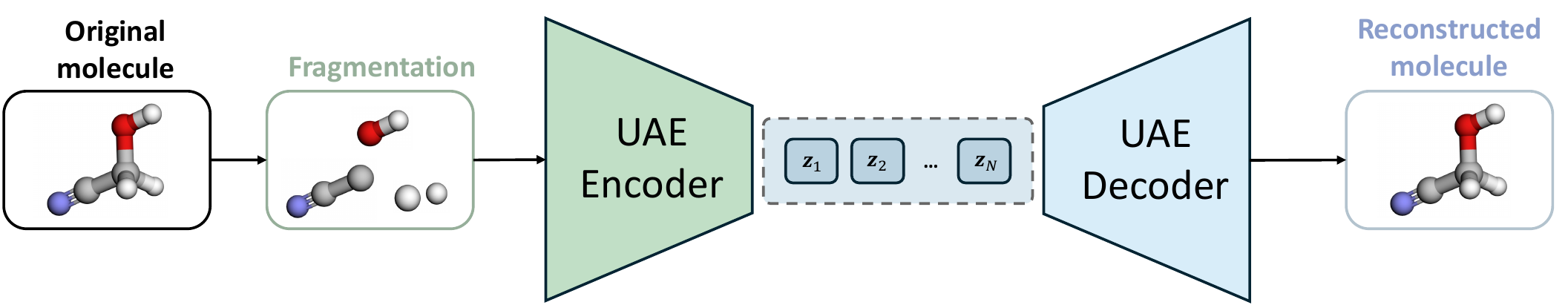}
    \caption{To enable fragment-conditioned generation, molecules are decomposed into disconnected fragment components before UAE encoding by removing inter-fragment bonds. Latent representations are therefore computed using only intra-fragment information. The UAE decoder subsequently reconstructs the full molecular graph, including bonds between fragments.}
    \label{fig:frag_local}
\end{figure}
In the original UAE architecture~\citep{luo2025towards}, latent representations are computed through message passing across the full molecular graph, allowing each latent token to aggregate information from progressively larger molecular neighborhoods. 
Consequently, the latent representation of each fragment depends not only on its internal structure but also on atoms belonging to neighboring fragments. However, our fragment-conditioned generation setting requires latent representations that can be computed for fragments in isolation at inference time. To achieve this, we modify the encoder connectivity during UAE training. Given a fragment decomposition $\{\mathcal F_0,\dots,\mathcal F_{K-1}\}$ of a molecular graph $\mathbf{G}$, we remove all inter-fragment edges prior to encoder message passing, restricting information propagation to atoms within the same fragment (Figure~\ref{fig:frag_local}). Importantly, this modification affects only the graph used by the encoder to compute latent representations. The decoder continues to reconstruct the complete molecule graph, including inter-fragment connectivity and global 3D geometry. Consequently, latent tokens are encouraged to capture fragment-local information, while the decoder integrates information across fragments and recovers global structure. Empirically, UAE reconstruction performance remains unchanged even when all inter-atomic edges are removed during encoding, suggesting that the decoder can recover global relationships from the collection of local latent representations. We attribute this flexibility to the decoder's joint attention over latent tokens and to the atom embeddings already encoding local identity and spatial information.

\subsection{Fragment statistics}\label{sec:fragment_stats}
\begin{figure}[!h]
    \centering
    %\includesvg[width=0.98\textwidth]{figures/grouped_ablation_comparison.svg}
    \includegraphics[width=0.98\columnwidth]{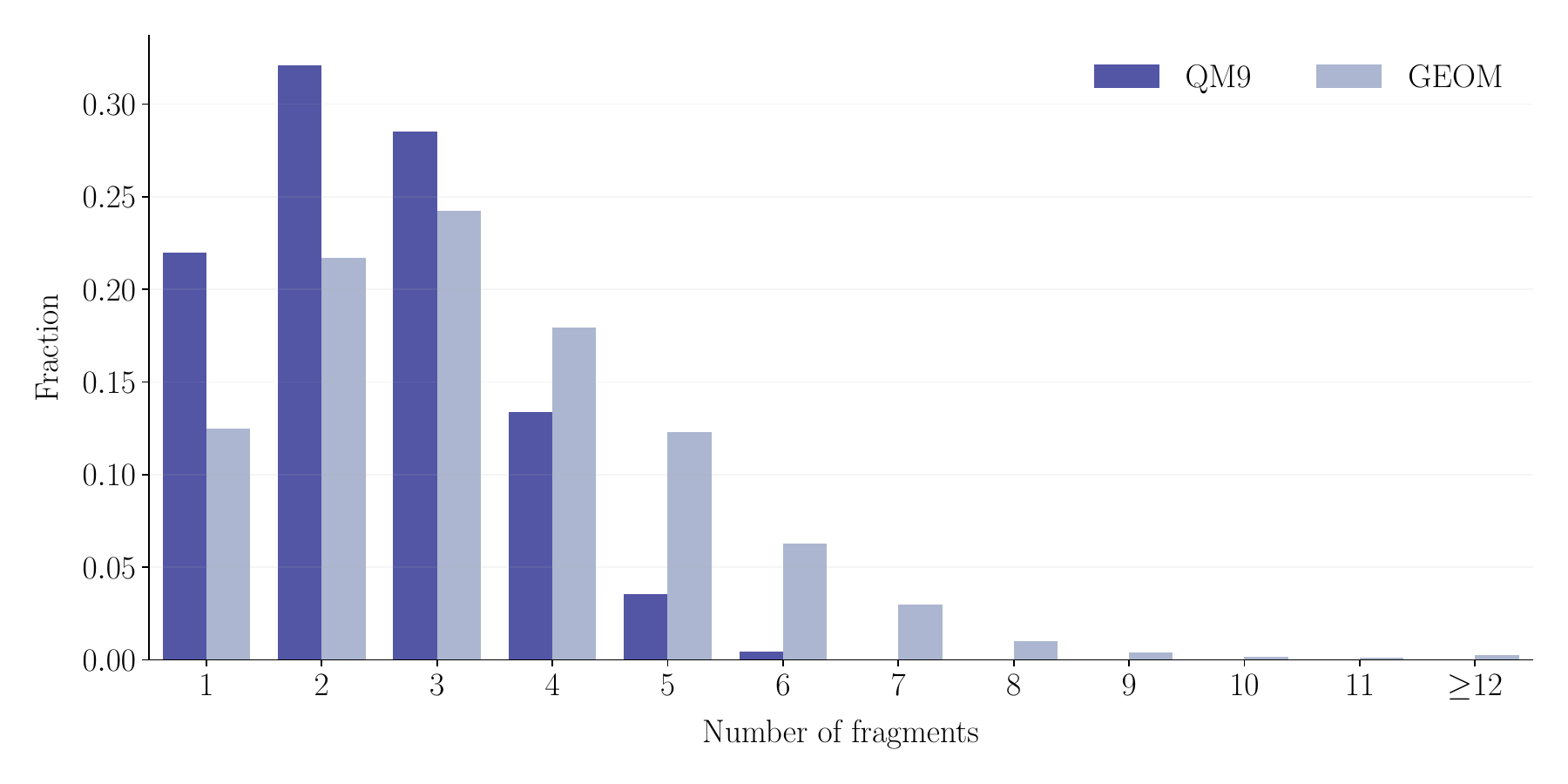}
    \caption{Distribution of molecular fragment counts across molecules in QM9 and GEOM.}
    \label{fig:frag_counts}
\end{figure}

Figure~\ref{fig:frag_counts} shows the distribution of BRICS fragment counts $K$ across molecules in QM9 and GEOM; the structural-hydrogen pseudo-fragment $\mathcal{F}_H$ is not included. QM9 molecules contain an average of $2.46$ fragments (median $=2$), while GEOM molecules exhibit greater structural diversity, averaging $3.40$ fragments (median $=3$).

\subsection{Evaluation Protocol}\label{app:eval_protocol}
All metrics are computed following the evaluation protocols used in prior 3D molecule generation work~\citep{luo2025towards}.

\paragraph{3D Structural Metrics.}
We first evaluate generated molecules directly from their predicted 3D coordinates. 2D molecular graphs are reconstructed from atomic coordinates via distance-based bond inference~\citep{hoogeboom2022ardm}, after which chemical validity and stability are evaluated on the reconstructed graphs.
\begin{itemize}

    \item \textbf{3D Validity}: fraction of generated conformations that can be converted into chemically-valid 2D molecular graphs.

    \item \textbf{3D AtomStable}: fraction of atoms satisfying valence constraints under the reconstructed 2D bond order.

    \item \textbf{3D MolStable}: fraction of molecules whose reconstructed 2D molecular graph satisfies chemical stability constraints.

\end{itemize}
%Generated latent sequences are decoded using the frozen UAE decoder to obtain molecular graphs and 3D coordinates. %All reported results
%are averaged over three independent runs.
%For unconditional generation, molecules are sampled autoregressively until the model predicts a STOP token or a predefined maximum sequence length is reached. For fragment-conditioned generation, the model is conditioned on a subset of molecular fragments and autoregressively generates the remaining latent tokens.
We also compare local geometric statistics between generated and reference conformations.
Specifically, we measure the discrepancy of bond length, bond angle, and dihedral angle distributions using \textit{Maximum Mean Discrepancy} (MMD). Lower values indicate closer agreement with the reference molecular geometry distribution and therefore more realistic 3D conformations.
\begin{itemize}
    \item \textbf{Bond Length}: MMD between generated and
    reference bond length distributions for eight frequent bond types (e.g., C--C, C--N, C--O). Computed using Gaussian kernel with bandwidth tuned via the median heuristics.
    \item \textbf{Bond Angle}: MMD between bond angle distributions of generated and reference molecules. Bond angles are defined over triplets of bonded atoms (e.g. C--C--C).
    \item \textbf{Dihedral Angle}: MMD between dihedral (torsional) angle distributions of generated and reference molecules. Dihedral angles are defined over bonded four-atom sequences (e.g. C--C--C--C).
\end{itemize}

\paragraph{2D Graph Metrics}

We additionally evaluate the quality of the explicitly predicted 2D molecular graph topology obtained directly from the decoder outputs, without relying on coordinate-based bond reconstruction. In this setting, validity and stability are evaluated directly on the predicted 2D molecular graphs.

\begin{itemize}
    \item \textbf{2D Validity}: fraction of predicted molecular graphs that satisfy chemical validity constraints.
    \item \textbf{2D AtomStable}: fraction of atoms satisfying valence constraints in the predicted 2D graph topology.
    \item \textbf{2D MolStable}: fraction of predicted 2D molecular graphs satisfying molecule-level chemical stability constraints.
    \item \textbf{Validity \& Completeness (V\&C)}: percentage of generated molecules that correspond to connected molecular graphs without disconnected fragments, and that can be converted into valid SMILES representations.%Fraction of fully connected, syntactically correct (SMILES-parsable) molecules, excluding fragmented or hypervalent structures.
    \item \textbf{Validity \& Uniqueness (V\&U)}: percentage of chemically valid generated molecules that are unique under canonical SMILES comparison. %Percentage of unique molecules among valid ones, calculated  as the ratio of unique SMILES strings to the total number of valid molecules.
    \item \textbf{Validity \& Uniqueness \& Novelty (V\&U\&N)}: percentage of generated molecules that are simultaneously chemically valid, unique, and absent from the training set under canonical SMILES comparison. %Percentage of unique molecules among valid  ones that are also novel, calculated as the ratio of unique SMILES strings to the total number of valid molecules, excluding those present in the training set.
\end{itemize}

\textbf{V\&C}, \textbf{V\&U}, \textbf{V\&U\&N} are evaluated only on directly predicted 2D molecular graphs, since uniqueness and novelty are defined with respect to 2D molecular identity (atom and bond topology). Different 3D conformations of the same 2D molecular graph are therefore not considered distinct or novel molecules.
\newpage
\subsection{Effect of Atom Ordering}
\label{app:perm-ablation}
Autoregressive molecular generators are sensitive to the order in which atoms are generated, as the chosen factorization determines the sequential dependencies learned by the model~\citep{cheng2025quetzal,haviv2022transformer,kazemnejad2023length}. Our fragment-based construction introduces variation at the level of molecular fragments while retaining a deterministic ordering of atoms within each fragment. To assess the importance of this structured ordering, we compare against training with fully random atom permutations. 
\begin{table}[!h]
\centering\small
\setlength{\tabcolsep}{6pt}
\caption{Effect of atom ordering on QM9 dataset. We report Mean $\pm$ std over three independent inference runs.}
\label{tab:perm-ablation}
\begin{tabular}{lccc}
\toprule
Atom ordering & 3D Validity$\uparrow$ & 3D AtomStable$\uparrow$ & 3D MolStable$\uparrow$ \\
& (\%) & (\%) & (\%) \\
\midrule
Fragment-based ($p_f{=}0.5$)     & 97.3$_{\pm0.1}$ & 99.07$_{\pm0.02}$ & 92.6$_{\pm0.1}$ \\
Default ($p_f{=}0.0$)     & 98.0$_{\pm0.1}$ & 99.34$_{\pm0.01}$ & 94.9$_{\pm0.0}$ \\
\midrule
Random permutation & 74.2$_{\pm0.4}$ & 88.40$_{\pm0.09}$ & 35.2$_{\pm0.2}$ \\
\bottomrule
\end{tabular}
\end{table}
As shown in Table~\ref{tab:perm-ablation}, random atom ordering substantially degrades unconditional generation: 3D Validity and 3D MolStable decrease from $98.0\%$ and $94.9\%$ under the default ordering to $74.2\%$ and $35.2\%$, respectively. In contrast, the mixed fragment-conditioned training strategy ($p_f = 0.5$) retains performance close to the default ordering, achieving $97.3\%$ 3D Validity and $92.6\%$ 3D MolStable. These results suggest that introducing diversity through structured fragment-level factorizations preserves useful sequential regularities, whereas arbitrary atom-level permutations substantially disrupt autoregressive generation.

\subsection{KRONOS Sampling Procedure}\label{app:sampling}
Algorithm~\ref{alg:sampling} describes sampling for both unconditional and fragment-conditioned generation modes; the two differ only in the prompt $P$. For unconditional generation $P = [\mathrm{START}]$ and $M = N_{\max}$, with $N_{\max}$ denoting the maximum number of allowed atoms. For fragment-conditioned generation $P = [\mathrm{START}] \circ [\mathrm{SEED}] \circ \mathbf{Z}_\mathcal{S}  \circ [\mathrm{GEN}]$, where $\mathbf{Z}_\mathcal{S}$ are the seed fragment latents, $\circ$ denotes sequence concatenation, and $M = N_{\max} - |\mathbf{Z}_\mathcal{S}|$; in the unconditional case $\mathbf{Z}_\mathcal{S} = ()$ and the returned molecule is decoded from the generated tokens alone.
\begin{algorithm}[h]
\caption{KRONOS sampling}
\begin{algorithmic}[1]
\Require transformer, denoiser $\varepsilon_\theta$, stop head $\mathrm{MLP}_{\mathrm{stop}}$,
         decoder $\mathcal{D}_{\mathrm{UAE}}$; prompt $P$, threshold $\tau$,
         DDIM steps $S$, maximum number of atoms $M$
\State $\mathbf{Z} \gets ()$
\For{$i = 1, \dots, M$}
    \State $\mathbf{c}_i \gets \mathrm{Transformer}\big(P \circ \mathbf{Z}\big)$ \Comment{context at the final position}
    \If{$\sigma(\mathrm{MLP}_{\mathrm{stop}}(\mathbf{c}_i)) > \tau$}
        \State \textbf{break} \Comment{terminate with $|\mathbf{Z}|$ generated atoms}
    \EndIf
    \State $\mathbf{z}_i \gets \mathrm{DDIM}(\varepsilon_\theta, \mathbf{c}_i, S)$ \Comment{$S$ denoiser evaluations}
    \State $\mathbf{Z} \gets \mathbf{Z} \circ \mathbf{z}_i$
\EndFor
\State \Return $\mathcal{D}_{\mathrm{UAE}}(\mathbf{Z}_\mathcal{S} \circ \mathbf{Z})$
\end{algorithmic}
\label{alg:sampling}
\end{algorithm}

\newpage
\subsection{Sampling efficiency} \label{app:runtime}
\begin{figure}[!h]
    \centering
    %\includesvg[width=0.98\textwidth]{figures/grouped_ablation_comparison.svg}
    \includegraphics[width=0.85\columnwidth]{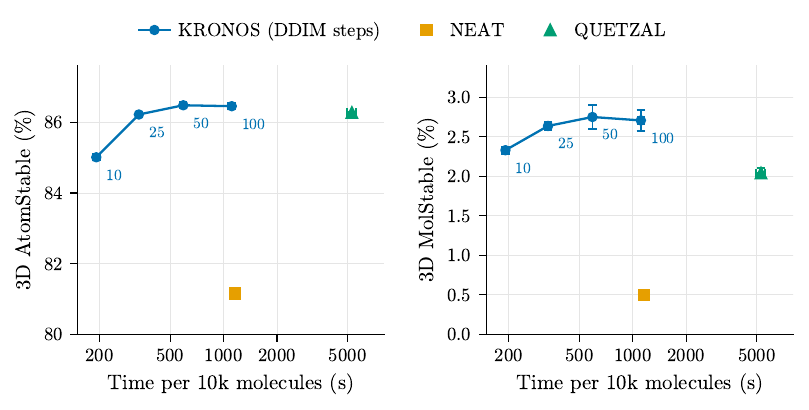}
    \caption{Quality versus sampling cost for $10$k molecules generated by each model trained on GEOM-Drugs dataset
    (mean $\pm$ std over three runs). KRONOS retains the highest 3D MolStable down to $25$ DDIM steps, at under a third of NEAT's sampling time.}
    \label{fig:runtime}
\end{figure}
We measure wall-clock time of KRONOS compared to two main autoregressive baselines (QUETZAL and NEAT). We sample $10$k GEOM-Drugs molecules on a single NVIDIA Quadro RTX 6000 (24 GB), averaged over three runs, each method using the largest batch size fitting in GPU memory
(KRONOS: $512$; QUETZAL: $2,048$; NEAT: $10,000$). Absolute times are not comparable to those reported by \citet{rose2025neat} on an RTX 4090, where
NEAT and QUETZAL are respectively $5.6\times$ and $4.8\times$ faster; the ordering between methods is
unchanged.

Sampling cost splits between a transformer backbone, run once per atom, and a denoiser, run for a number of steps per atom. As sampler order differs, we report denoiser network function evaluations (NFE) alongside the number of steps each method is configured with: KRONOS uses first-order DDIM~\citep{song2021denoising} with $100$ steps, NEAT employs first-order Euler--Maruyama with $60$ steps (NFE = steps), whereas QUETZAL uses a second-order Heun sampler~\citep{karras2022elucidating} with $120$ steps (NFE = $2 \cdot \text{steps} -1$), giving $4,400$, $2,640$ and $10,516$ evaluations for an average 44-atom molecule in GEOM-Drugs. The three denoisers share width and depth (1,536 channels, 6 residual blocks), so per-evaluation cost is comparable.

Backbone cost differs structurally. KRONOS attends causally, so an atom's representation is unaffected
by atoms placed after it and is retained in a key--value (KV) cache: one backbone pass per atom. NEAT attends
bidirectionally over the atom set, so cached state cannot be reused and the partial molecule is re-encoded at every step. For a $44$-atom molecule this amounts to $44$ backbone passes for KRONOS against $990$ for NEAT, offsetting NEAT's lower number of denoiser evaluations. QUETZAL is causal and could in principle cache, but its released implementation re-encodes the sequence at each step; even with caching, its second-order sampler at 120 steps requires $239$ denoiser evaluations per atom against KRONOS's $100$.

At its default setting of 100 DDIM steps, KRONOS samples in $1110 \pm 47$~s, comparable to NEAT ($1156 \pm 77$~s) and 4.8$\times$ faster than QUETZAL ($5273 \pm 309$~s). Figure~\ref{fig:runtime} reports quality against sampling cost for all three. Fewer steps scale denoiser work down proportionally while backbone cost is fixed by the atom count: at 25 steps KRONOS takes
$333 \pm 6$~s, 3.5$\times$ faster than NEAT and 15.9$\times$ faster than QUETZAL, with stability essentially unchanged (3D AtomStable $86.2$ vs $86.5$ (\%); 3D MolStable $2.64$ vs $2.71$ (\%)).

\newpage
\subsection{2D molecule generation}\label{app:2d_gen}
Because KRONOS operates in the unified latent space of the UAE, generated latent tokens can be decoded into both 3D molecular structures and explicit 2D molecular graphs (utilizing predicted bond topology). This enables evaluation of the corresponding 2D molecules and direct comparison with prior methods operating on mixed 2D/3D molecular representations~\citep{vignac2023midi, huang2024jodo, irwin2024semlaflow}. We evaluate $10$k generated molecules per model and report the resulting 2D metrics in Table~\ref{tab:2d_gen}.
Overall, KRONOS achieves competitive 2D generation performance across both benchmarks. While SemlaFlow attains the strongest overall 2D metrics, KRONOS consistently generates highly valid and stable molecular graphs.
%These findings indicate that the proposed latent autoregressive diffusion process preserves the global information encoded in the latent representation, enabling high-quality generation in both 2D and 3D molecular modalities.
\begin{table*}[!h]
\centering
\small
\caption{
2D molecule generation metrics. We report metrics on the test sets of the corresponding datasets in \textcolor{gray}{gray}. Models marked with (*) were evaluated using released generated molecules.
Models marked with (**) were evaluated using official pre-trained weights and source code. \textbf{Best} and \underline{second-best} entries are highlighted in bold and underlined, respectively. We report the $\text{mean} \pm \text{std}$ over three independent runs.
}
\label{tab:2d_gen}
\vspace{-2mm}
\resizebox{\textwidth}{!}{
\begin{tabular}{lrrrrrrr}
\toprule
\textbf{Model}
& \multicolumn{1}{c}{\textbf{2D Validity} ($\uparrow$)}
& \multicolumn{1}{c}{\textbf{2D AtomStable} ($\uparrow$)}
& \multicolumn{1}{c}{\textbf{2D MolStable} ($\uparrow$)}
& \multicolumn{1}{c}{\textbf{V\&C} ($\uparrow$)}
& \multicolumn{1}{c}{\textbf{V\&U} ($\uparrow$)}
& \multicolumn{1}{c}{\textbf{V\&U\&N} ($\uparrow$)} \\

& \multicolumn{1}{c}{(\%)}
& \multicolumn{1}{c}{(\%)}
& \multicolumn{1}{c}{(\%)}
& \multicolumn{1}{c}{(\%)}
& \multicolumn{1}{c}{(\%)}
& \multicolumn{1}{c}{(\%)} \\
\midrule

%\multicolumn{7}{c}{\textbf{QM9}} \\
\textcolor{gray}{QM9}
& \textcolor{gray}{$100.0\phantom{_{\pm 0.0}}$} & \textcolor{gray}{$100.00\phantom{_{\pm 0.00}}$} & \textcolor{gray}{$100.00\phantom{_{\pm 0.00}}$} & \textcolor{gray}{$100.0\phantom{_{\pm 0.0}}$} & \textcolor{gray}{$99.9\phantom{_{\pm 0.0}}$} & \textcolor{gray}{$99.9\phantom{_{\pm 0.0}}$} \\
\midrule
%UDM-3D
%& $93.0_{\pm 0.7}$ & $99.60_{\pm 0.04}$ & $93.40_{\pm 0.70}$ & $93.0_{\pm 0.7}$ & $90.6_{\pm 0.9}$ & $90.6_{\pm 0.9}$ \\
\multicolumn{7}{l}{\textit{Diffusion / Flow}}\\
\midrule
JODO*
& $98.1\phantom{_{\pm 0.0}}$ & $99.83\phantom{_{\pm 0.00}}$ & $97.83\phantom{_{\pm 0.00}}$ & $98.1\phantom{_{\pm 0.0}}$ & $\mathbf{95.1}\phantom{_{\pm 0.0}}$ & $\mathbf{95.1}\phantom{_{\pm 0.0}}$ \\

MiDi**
& $96.8_{\pm0.1}$ & $99.74_{\pm0.01}$ & $96.33_{\pm0.13}$ & $96.7_{\pm0.1}$ & $94.4_{\pm0.1}$ & $94.3_{\pm0.1}$ \\

SemlaFlow**
& $\mathbf{99.4}_{\pm0.1}$ & $\mathbf{100.00}_{\pm0.00}$ & $\mathbf{99.63}_{\pm0.06}$ & $\mathbf{99.3}_{\pm0.1}$ & $\underline{94.8}_{\pm0.1}$ & $\underline{94.8}_{\pm0.1}$ \\

\midrule
\multicolumn{7}{l}{\textit{Autoregressive}}\\
\addlinespace[2pt]
\rowcolor{blue!15}
\textbf{KRONOS (ours)}
& $\underline{99.2}_{\pm 0.2}$ & $\underline{99.92}_{\pm 0.02}$ & $\underline{99.10}_{\pm 0.02}$ & $\underline{99.1}_{\pm 0.2}$ & $92.5_{\pm 0.6}$ & $92.5_{\pm 0.6}$ \\
\bottomrule
\addlinespace[20pt]
\midrule
%\multicolumn{7}{c}{\textbf{GEOM}} \\
\textcolor{gray}{GEOM}
& \textcolor{gray}{$89.4\phantom{_{\pm 0.0}}$} & \textcolor{gray}{$99.44\phantom{_{\pm 0.00}}$} & \textcolor{gray}{$86.27\phantom{_{\pm 0.00}}$} & \textcolor{gray}{$89.4\phantom{_{\pm 0.0}}$} & \textcolor{gray}{$18.5\phantom{_{\pm 0.0}}$} & \textcolor{gray}{$18.5\phantom{_{\pm 0.0}}$} \\
\midrule

%UDM-3D
%&  &  & &  & & \\
\multicolumn{7}{l}{\textit{Diffusion / Flow}}\\
\midrule
JODO*
& $79.1\phantom{_{\pm 0.0}}$ & $\underline{99.26}\phantom{_{\pm 0.00}}$ & $80.39\phantom{_{\pm 0.00}}$ & $75.6\phantom{_{\pm 0.0}}$ & $78.1\phantom{_{\pm 0.0}}$ & $78.1\phantom{_{\pm 0.0}}$  \\

MiDi**
& $59.9_{\pm0.2}$ & $96.43_{\pm0.11}$ & $72.85_{\pm0.32}$ & $57.9_{\pm0.2}$ & $59.9_{\pm0.2}$ & $59.9_{\pm0.2}$ \\

SemlaFlow**
& $\mathbf{95.1}_{\pm0.1}$ & $\mathbf{100.00}_{\pm0.00}$ & $\mathbf{98.63}_{\pm0.11}$ & $\mathbf{92.4}_{\pm0.2}$ & $\mathbf{95.1}_{\pm0.2}$ & $\mathbf{95.1}_{\pm0.2}$ \\

\midrule
\multicolumn{7}{l}{\textit{Autoregressive}}\\
\addlinespace[2pt]
\rowcolor{blue!15}
\textbf{KRONOS (ours)}
& $\underline{84.6}_{\pm 0.4}$ & $99.16_{\pm 0.02}$ & $\underline{83.45}_{\pm 0.29}$ & $\underline{84.4}_{\pm0.4}$ & $\underline{84.5}_{\pm0.5}$ & $\underline{84.5}_{\pm0.5}$ \\

\bottomrule
\end{tabular}
}
\label{table:metrics_2d}

\end{table*}
\newpage
\subsection{Scaling ablations}
Tables~\ref{tab:qm9_scaling_abl} and~\ref{tab:geom_scaling_abl} evaluate the effect of scaling the transformer backbone while keeping the diffusion head fixed at dimension $1536$. We vary the number of layers, attention heads, and hidden dimension. Across both QM9 and GEOM-Drugs, increasing model capacity consistently improves or maintains generation performance. %with the largest model achieving the strongest overall results. These findings suggest that KRONOS scales favorably with Transformer capacity.
\begin{table}[h]
\centering
\small
\setlength{\tabcolsep}{4pt}
\caption{Model ablations results on the QM9 dataset. We report $\text{mean} \pm \text{std}$ across three independent runs. Best in \textbf{bold}, second-best \underline{underlined}. The best configuration is highlighted in \textcolor{blue!60}{blue}.}
\label{tab:qm9_scaling_abl}

\begin{tabular}{ccccccc}
\toprule
\begin{tabular}[c]{@{}c@{}}Transformer\\ heads\end{tabular} &
\begin{tabular}[c]{@{}c@{}}Transformer\\ layers\end{tabular} &
\begin{tabular}[c]{@{}c@{}}Transformer\\ dimension\end{tabular} &
\begin{tabular}[c]{@{}c@{}}Model parameters\\ $(\times 10^6)$\end{tabular} &
\begin{tabular}[c]{@{}c@{}}3D\\ Validity\end{tabular} &
\begin{tabular}[c]{@{}c@{}}3D\\ AtomStable\end{tabular} &
\begin{tabular}[c]{@{}c@{}}3D\\ MolStable\end{tabular} \\
\midrule
%6  & 6  & 384 & 89.46  & 91.5 & 96.8 & 74.8 \\
$8$  & $8$  & $512$ & $105.27$ & $95.17_{\pm 0.04}$ & $98.55_{\pm 0.01}$ & $88.7_{\pm 0.2}$ \\
$10$ & $10$ & $640$ & $129.61$ & $\underline{95.90}_{\pm 0.11}$ & $98.67_{\pm 0.04}$ & $89.5_{\pm 0.3}$ \\
$12$ & $12$ & $768$ & $165.78$ &
$95.87_{\pm 0.08}$ & $\underline{98.73}_{\pm 0.01}$ & $\underline{90.1}_{\pm 0.1}$ \\
\rowcolor{blue!15} $16$ & $16$ & $1024$ & $283.05$ & $\mathbf{97.29}_{\pm 0.13}$& $\mathbf{99.07}_{\pm 0.02}$ & $\mathbf{92.6}_{\pm 0.1}$  \\
\bottomrule
\end{tabular}
\end{table}

\begin{table}[h]
\centering
\small
\setlength{\tabcolsep}{4pt}
\caption{Model scaling ablations results on the GEOM-Drugs dataset. We report $\text{mean} \pm \text{std}$ across three independent runs. Best in \textbf{bold}, second-best \underline{underlined}. The best configuration is highlighted in \textcolor{blue!60}{blue}.}
\label{tab:geom_scaling_abl}

\begin{tabular}{ccccccc}
\toprule
\begin{tabular}[c]{@{}c@{}}Transformer\\ heads\end{tabular} &
\begin{tabular}[c]{@{}c@{}}Transformer\\ layers\end{tabular} &
\begin{tabular}[c]{@{}c@{}}Transformer\\ dimension\end{tabular} &
\begin{tabular}[c]{@{}c@{}}Model parameters\\ $(\times 10^6)$\end{tabular} &
\begin{tabular}[c]{@{}c@{}}3D\\ Validity\end{tabular} &
\begin{tabular}[c]{@{}c@{}}3D\\ AtomStable\end{tabular} &
\begin{tabular}[c]{@{}c@{}}3D\\ MolStable\end{tabular} \\
\midrule
%6  & 6  & 384 & 89.46  & 91.5 & 96.8 & 74.8 \\
$8$  & $8$  & $512$ & $105.35$ & $93.2_{\pm 0.1}$ & $86.42_{\pm 0.11}$ & $\mathbf{2.73}_{\pm 0.08}$ \\
$10$ & $10$ & $640$ & $129.71$ & $93.1_{\pm 0.2}$ & $86.23_{\pm 0.05}$ & $2.69_{\pm 0.05}$ \\
$12$ & $12$ & $768$ & $165.90$ &
$\underline{93.4}_{\pm 0.2}$ & $\underline{86.43}_{\pm 0.06}$ & $2.64_{\pm 0.11}$ \\
\rowcolor{blue!15} $16$ & $16$ & $1024$ & $283.21$ & $\mathbf{94.7}_{\pm 0.1}$ & $\mathbf{86.45}_{\pm0.03}$ & $\underline{2.71}_{\pm0.13}$  \\
\bottomrule
\end{tabular}
\end{table}
\subsection{Effect of fragment-conditioning probability $p_f$}\label{app:p_f}
\begin{figure}[!h]%{r}{0.46\textwidth}
\centering
\includegraphics[width=0.65\textwidth]{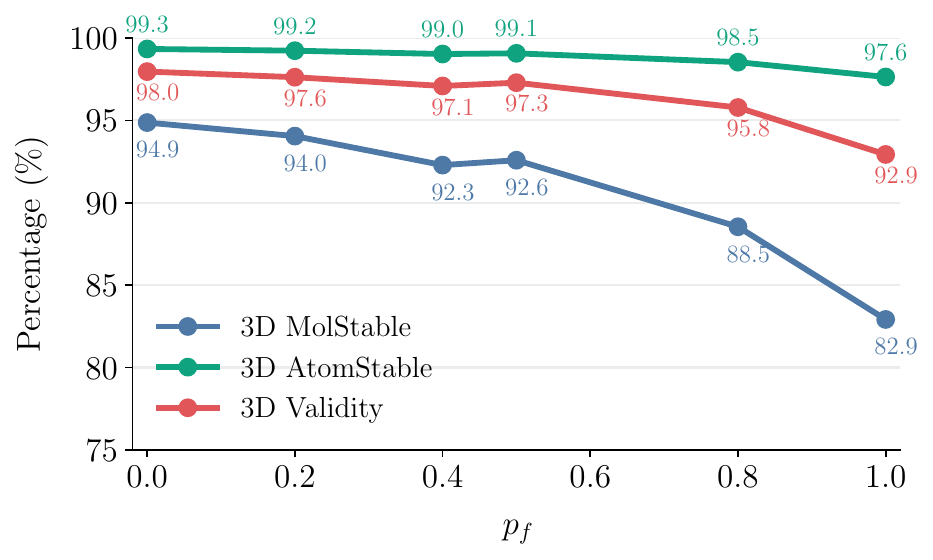}
\caption{
Effect of the fragment-conditioning probability $p_f$ on unconditional molecule generation performance (QM9). Increasing $p_f$ gradually trades unconditional generation quality for fragment-conditioning capability.
}
\label{fig:p_f}
\end{figure}
During training, each molecule is transformed to a fragment-conditioned sequence with probability $p_f$. Figure~\ref{fig:p_f} reports unconditional generation performance on QM9 as $p_f$ varies. We observe that model's performance on unconditional generation remains stable for moderate values of $p_f$, with performance degrading only as training becomes dominated by fragment-conditioned examples. Based on these observations, we select $p_f=0.5$ throughout the main experiments as a reasonable trade-off between unconditional and fragment-conditioned generation.

\subsection{Generated molecule size distribution}
\begin{figure}[!h]%{r}{0.46\textwidth}
\centering
\includegraphics[width=0.98\textwidth]{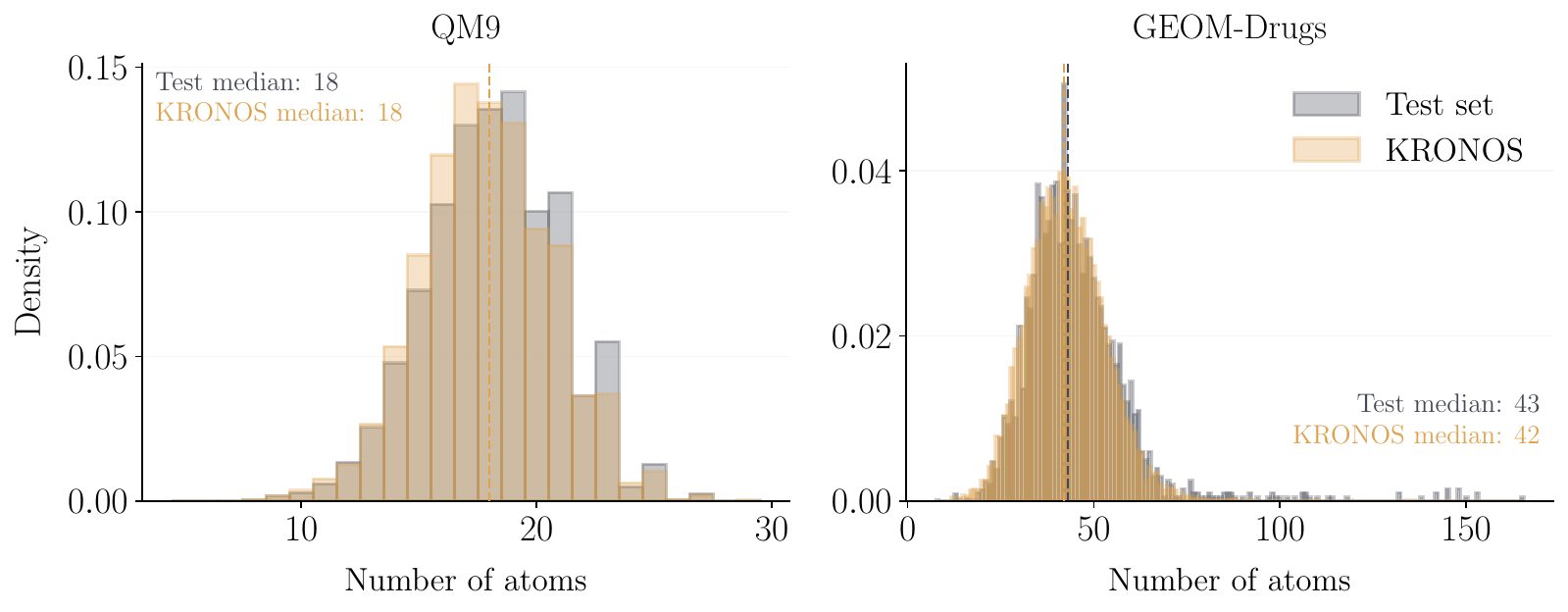}
\caption{Distribution of the number of atoms in molecules generated by KRONOS and in the corresponding test sets for QM9 and GEOM-Drugs. KRONOS closely reproduces the molecule size distributions of both datasets.}
\label{fig:atom_counts}
\end{figure}

Figure~\ref{fig:atom_counts} compares the distribution of the number of atoms in $10$k generated molecules against $10$k molecules sampled from the corresponding test sets. On QM9, the generated and test distributions exhibit identical median molecule sizes (18 atoms), while on GEOM-Drugs the generated molecules have a median of 42 atoms compared to 43 atoms in the test set. Overall, the learned stopping criterion produces molecule-size distributions that closely match those observed in the data, indicating that KRONOS effectively learns when to terminate generation.

\subsection{Limitations and future work} \label{app:lim_and_future}
KRONOS inherits limitations from the underlying latent autoregressive formulation.
First, generation quality is bounded by the reconstruction quality of a separately trained autoencoder~\citep{luo2025towards}. Improvements in latent representation learning may therefore translate directly into improved generative performance.

Second, while fragment-conditioned training substantially improves completion from partial molecular context, KRONOS is not permutation-invariant. \citet{rose2025neat} obtain this property by sum pooling atom representations into a single vector, which they note may cause oversquashing. Future work may find better trade-offs between expressivity and permutation invariance.

Third, conditioning fragments are supplied with the atomic coordinates they occupy in a reference conformer, and KRONOS is sensitive to this placement: fragments positioned outside the distribution seen during training yield shorter and less reliable completions. We found that discarding absolute positions in favour of relative geometry alone is not a straightforward remedy: in our early experiments, encoding distance-based features instead of absolute coordinates degraded UAE reconstruction considerably. A potential direction, which we leave to future work, is to supply the missing placement rather than remove the dependence on it. A fragment specified without coordinates can be embedded with a standard conformer generator, and positioned by sampling a placement observed for a training fragment of comparable size. This would keep the encoder input in-distribution and enable conditioning on a fragment vocabulary without a reference molecule.

Fourth, our fragment decomposition relies on a fixed BRICS-derived partitioning scheme. While this provides chemically meaningful conditioning units, alternative decomposition strategies may expose different structural regularities, and learning the decomposition jointly with the generative model remains an open direction.

Finally, while we focus on unconditional and fragment-conditioned molecule generation, the proposed framework could be naturally extended to property-guided molecular design. In particular, future work could condition generation on a molecular fragment while simultaneously optimizing for desired molecular properties.

\newpage
\subsection{Hyperparameters}
Tables~\ref{tab:qm9_hyperparameters} and \ref{tab:geom_hyperparameters} report the hyperparameters used to train KRONOS on QM9 and GEOM-Drugs, respectively.
\begin{table}[!h]
\centering
\caption{Hyperparameters of KRONOS trained on the QM9 dataset.}
\label{tab:qm9_hyperparameters}
\begin{tabular}{ll}
\toprule
\textbf{Parameter} & \textbf{Value} \\
\midrule
batch size & 2048 \\
bias & false \\
dropout & 0.0 \\
epochs & 10,000 \\
diffusion head hidden dimension & 1,536 \\
diffusion head layers & 6 \\
diffusion noise schedule & cosine \\
number of diffusion steps & 100 \\
sampler & DDIM \\
learning rate, weight decay & $1.0 \times 10^{-4}$, $10^{-6}$ \\
learning rate decay number of epochs (cosine annealing scheduling) & 10,000 \\
learning rate minimum (cosine annealing scheduling) & 10\% \\
learning rate warm-up number of epochs (cosine annealing scheduling) & 50 \\
time step resampling & 4 \\
transformer heads & 16 \\
stop classifier threshold $\tau$ & 0.5 \\
transformer hidden dimension & 1024 \\
transformer layers & 16 \\
stop classifier weight $\lambda_{stop}$ & 1.0 \\
fragment conditioning probability $p_f$ & 0.5 \\

\bottomrule
\end{tabular}
\end{table}
\begin{table}[!h]
\centering
\caption{Hyperparameters of KRONOS trained on the GEOM-Drugs dataset.}
\label{tab:geom_hyperparameters}
\begin{tabular}{ll}
\toprule
\textbf{Parameter} & \textbf{Value} \\
\midrule
batch size & 512 \\
bias & false \\
dropout & 0.0 \\
epochs & 1,000 \\
diffusion head hidden dimension & 1,536 \\
diffusion head layers & 6 \\
diffusion noise schedule & cosine \\
number of diffusion steps & 100 \\
sampler & DDIM \\
learning rate, weight decay & $1.0 \times 10^{-4}$, $10^{-6}$ \\
learning rate decay number of epochs (cosine annealing scheduling) & 1,000 \\
learning rate minimum (cosine annealing scheduling) & 10\% \\
learning rate warm-up number of epochs (cosine annealing scheduling) & 10 \\
time step resampling & 4 \\
transformer heads & 16 \\
transformer hidden dimension & 1024 \\
transformer layers & 16 \\
stop classifier threshold $\tau$ & 0.5 \\
stop classifier weight $\lambda_{stop}$ & 1.0 \\
fragment conditioning probability $p_f$ & 0.5 \\

\bottomrule
\end{tabular}
\end{table}
\newpage
\subsection{Generated molecules}\label{app:gen_mols}
\begin{figure}[!h]%{r}{0.46\textwidth}
\centering
\vspace{1cm}
\includegraphics[width=0.98\textwidth]{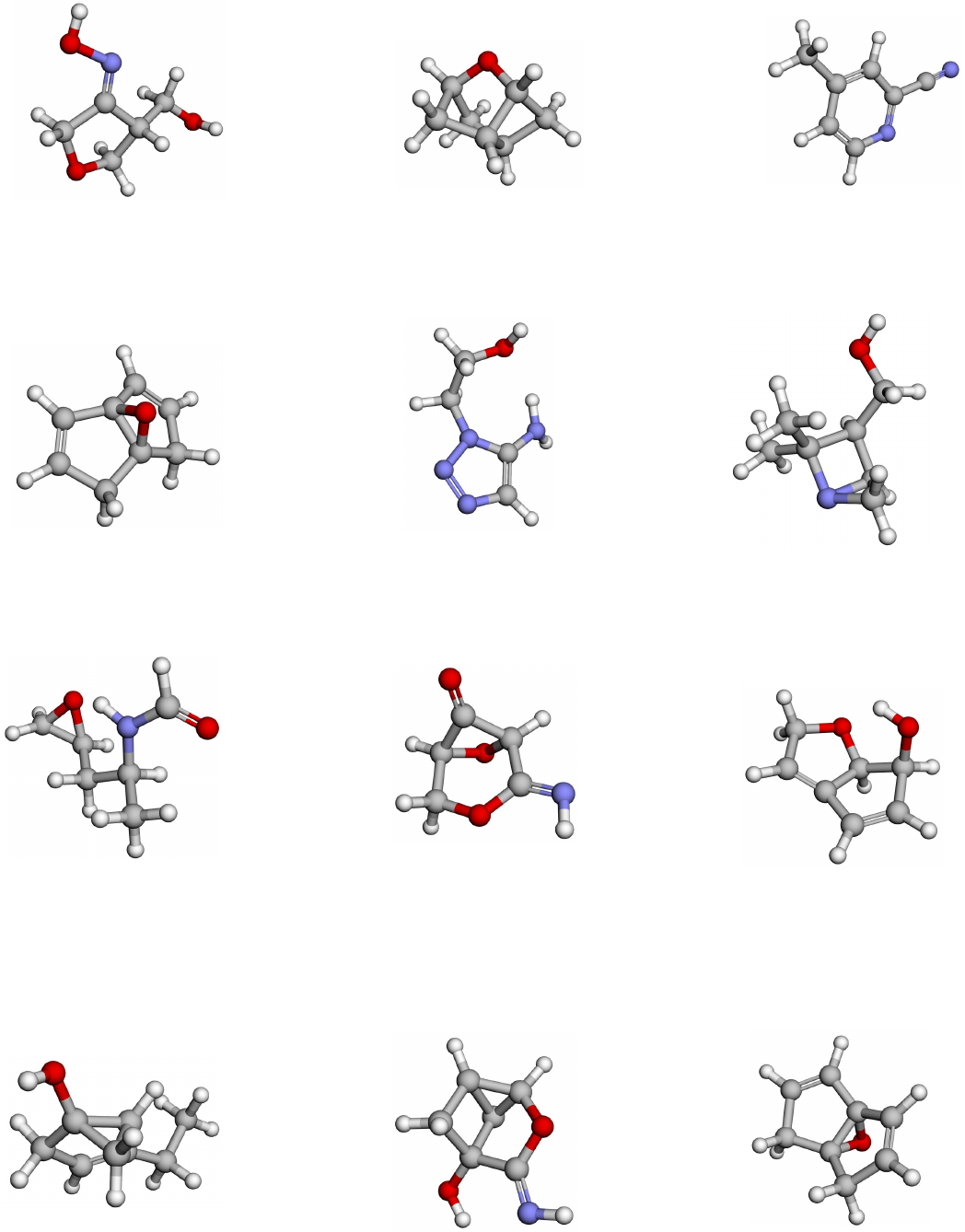}
\vspace{1cm}
\caption{Examples of 3D molecules generated by KRONOS trained on QM9 dataset.}
\label{fig:mols_qm9}
\end{figure}
\newpage
\begin{figure}[!h]
\centering
\vspace{2cm}
\includegraphics[width=0.98\textwidth]{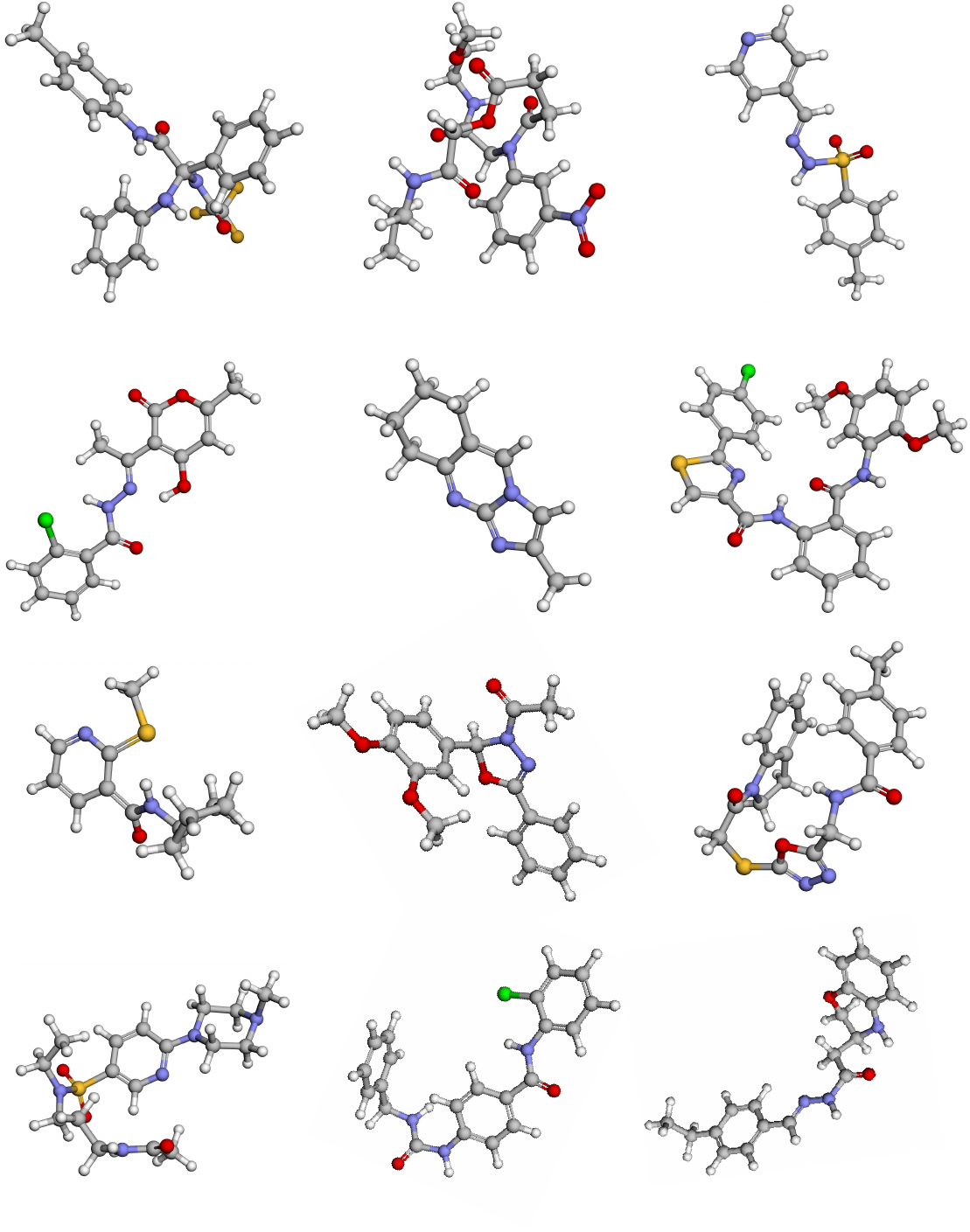}
\vspace{1cm}
\caption{Examples of 3D molecules generated by KRONOS trained on GEOM-Drugs dataset.}
\label{fig:mols_geom}
\end{figure}
\newpage
\begin{figure}[!h]
\centering
\includegraphics[width=0.92\textwidth]{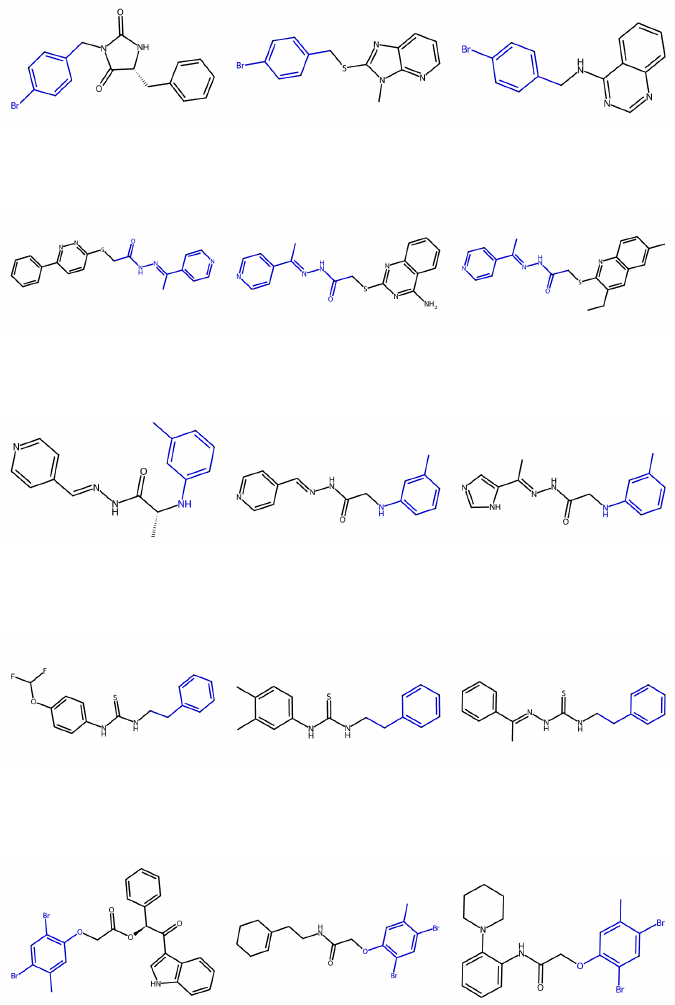}
\vspace{1cm}
\caption{Fragment-conditioned completions generated by KRONOS on GEOM-Drugs test molecules. Each row shows three
independent completions of the same conditioning fragment; the prefix fragment is shown in blue.}
\label{fig:prefix_completion_examples}
\end{figure}
\end{document}